\documentclass[11pt,a4paper]{article}
\usepackage[margin=2.5cm]{geometry}
\usepackage{amsmath,amssymb}
\usepackage{textcomp}
\usepackage[numbers,sort&compress]{natbib}
\usepackage{nameref}
\usepackage[hidelinks]{hyperref}
\usepackage[nopatch=eqnum]{microtype}
\usepackage[table]{xcolor}
\usepackage{soul}
\usepackage{booktabs}
\usepackage{array}
\usepackage{caption}
\usepackage{graphicx}
\usepackage{authblk}

\begin{document}
\title{Scientific Domain Knowledge Improves Vision-Language Fundus Models}

\author[1,2]{Verena Jasmin Hallitschke\thanks{Corresponding author: \texttt{verena-jasmin.hallitschke@uni-tuebingen.de}}}
\author[3]{Carsten Eickhoff}
\author[1,2]{Philipp Berens}

\affil[1]{Hertie Institute for AI in Brain Health, University of T\"ubingen, T\"ubingen, Germany}
\affil[2]{T\"ubingen AI Center, T\"ubingen, Germany}
\affil[3]{Institute for Applied Medical Informatics, University of T\"ubingen, T\"ubingen, Germany}

\date{\today}

\maketitle

\begin{abstract}
Vision-language models hold considerable promise for ophthalmology, but it remains unclear which training data source best conveys expert domain knowledge.
Existing ophthalmic models are trained on fixed text templates, medical reports, or general biomedical literature, sources that have never been compared under matched conditions.
To include domain-specific literature in this comparison, we present PubMed-Ophtha, a hierarchical dataset with high domain density of 102,023 panels with their subcaptions from 15,842 open-access articles in PubMed Central.
We then finetuned identical CLIP models on each source, using a general biomedical literature model as baseline, and found that domain-specific literature achieved the best average performance across 110 clinical tasks, reaching a mean linear probing AUROC of 88.63\% ahead of medical reports (85.68\%).
Restricting the dataset to fundus images, to the image count of the medical report dataset, or to articles unrelated to the evaluation datasets did not reduce performance, indicating that the gains likely stem from domain density.
We release the dataset, the finetuned models, and the full generation pipeline.
\end{abstract}

\section{Introduction}

\begin{figure}[t!]
    \centering
    \includegraphics[width=0.88\linewidth]{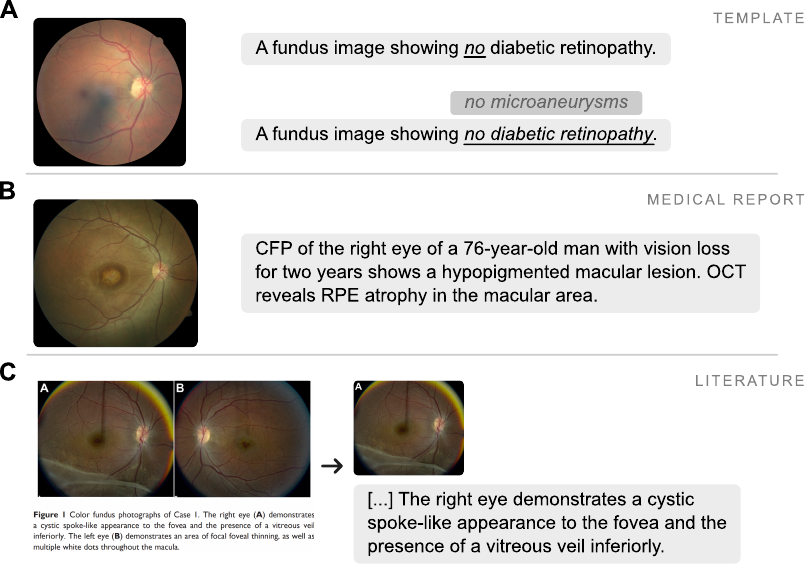}
    \caption{Overview of data sources and text prompts for training ophthalmic CLIP-based VLMs. (A) Simplistic and augmented templates offer high consistency but lack clinical nuance; here, images from the Kaggle EyePACS dataset are captioned by inserting the severity into a fixed template (top).
    In the FLAIR dataset, the categorical labels of several fundus datasets are additionally enriched with synonyms to increase linguistic diversity (bottom). (B) Medical reports provide real-world relevance but are often noisy and brief due to time constraints (example from the DeepEyeNet dataset). (C) Scientific literature provides curated, peer-reviewed knowledge; here, a figure from an ophthalmic article is decomposed into panels with their matching subcaptions, as in PubMed-Ophtha (example from~\cite{leng2012two}).
    General biomedical literature datasets contain the same text type, but cover all biomedical domains and therefore have a much lower domain density.}
    \label{fig:dataset-types-examples}
\end{figure}

After early deep learning approaches in ophthalmology had focused on image classification for disease grading, recent works in vision-language models (VLMs) integrate visual and textual encoders, facilitating automated clinical report generation and nuanced decision support \cite{lim2024vision}.
One of the most popular VLM architectures also in this domain has been Contrastive Language-Image Pre-Training (CLIP)~\cite{radford2021LearningTransferableVisuala}, which can be used for a wide range of tasks such as visual feature extraction and text-to-image retrieval~\cite{caffagni2024RevolutionMultimodalLarge}.
In fact, many multimodal large language models such as LLaVA~\cite{liu2023VisualInstructionTuninga} use CLIP-based models as visual feature extractors.
Many pretrained CLIP models rely on existing image-text pairs from the internet \cite{zhai2023sigmoid, sun2023eva} and have been adapted to medical data based on image-text pairs extracted from the biomedical literature, as in BiomedCLIP \cite{zhang2025MultimodalBiomedicalFoundation}.
VLMs for specific medical disciplines such as ophthalmology are then typically finetuned versions of these pretrained CLIP models~\cite{shi2025multimodal, du2024RETCLIPRetinalImage, CLIPDRTextualKnowledgeGuided}.
These models are trained using a range of paired image-text datasets, including synthetically generated paired datasets created from public image collections using text templates \cite{CLIPDRTextualKnowledgeGuided, silva2025foundation},
private clinical datasets containing medical reports with images \cite{shi2025multimodal} or rare and comparatively small public paired datasets containing medical reports \cite{huang2021DeepOphtMedicalReport}  (Fig.~\ref{fig:dataset-types-examples}).

Template-based datasets typically use simple sentence templates to turn existing categorical labels into descriptive prompts  such as CLIP-DR~\cite{CLIPDRTextualKnowledgeGuided} (Fig.~\ref{fig:dataset-types-examples}A).
Methods like FLAIR \cite{silva2025foundation} enhance this by incorporating synonyms to enrich categorical labels.
While approaches have proven highly effective for common disease types, including in ophthalmology~\cite{silva2025foundation} as they benefit from large-scale, expert-validated classification datasets, they remain constrained by their rigid linguistic structures.
The second type leverages medical reports \cite{huang2021DeepOphtMedicalReport}, which offer high clinical diversity and descriptive depth (Fig.~\ref{fig:dataset-types-examples}B), but these datasets are few and small: annotation requires clinical expertise, data privacy constraints limit scale, and the reports themselves can be inconsistent owing to the time pressure under which they are produced~\cite{reiner2007RadiologyReportingPresent,clynch2015MedicalDocumentationPart}.
Finally, medical literature datasets (e.g., BIOMEDICA \cite{lozanoBIOMEDICAOpenBiomedical}) utilize automated pipelines to extract figures, captions, and in-text mentions from peer-reviewed scientific publications in open-access repositories (Fig.~\ref{fig:dataset-types-examples}C).
Such corpora provide unparalleled coverage of rare diseases and evolving imaging modalities.

These datasets, however, share several limitations: many are not publicly available~\cite{qin2026VOLMOVersatileOpen, zhang2025MultimodalBiomedicalFoundation}, most are not specific to ophthalmology~\cite{baghbanzadehAdvancingMedicalRepresentationa, lozanoBIOMEDICAOpenBiomedical, zhang2025MultimodalBiomedicalFoundation, pelka2018roco, ruckert2024rocov2}, and all rely on the pre-extracted figure images provided in PubMed Central article packages, which are compressed and therefore of lower resolution than the original images.
Beyond image quality, none of these datasets account for the hierarchical structure of article figures — which are typically composed of labeled panels, each potentially containing multiple images.
Methods like MM-Retinal~\cite{wu2024MMRetinalKnowledgeEnhancedFoundational} and RetiZero~\cite{wang2025EnhancingDiagnosticAccuracy} use a mixture of template-based, medical report, and medical literature datasets.
Notably, the literature processing pipeline in both cases relies on expert supervision, which is not scalable to larger literature corpora. 

As existing ophthalmic CLIP models differ in model architecture, training schedule, and training data simultaneously, the contribution of the training data source alone cannot be isolated.
The trade-off between noisy clinical reports, simplistic templates, and curated scientific literature (Fig.~\ref{fig:dataset-types-examples}) therefore remains an under-explored direction.
To test the effect of literature datasets on ophthalmology specifically, a high-quality dataset with high domain density is required.
To include domain-specific literature in this comparison, we present PubMed-Ophtha, a large-scale, hierarchical dataset of ophthalmic panels with their subcaptions that directly addresses these limitations.
It contains 102,023 panels extracted from 15,842 open-access articles.
Unlike existing datasets, figures are extracted directly from article PDFs at full resolution and decomposed into their hierarchical structure of panels, panel identifiers, and individual images.
Our dataset represents the curated, peer-reviewed state of ophthalmic knowledge with much higher domain density and higher quality images than approaches based on the entire PubMed Central corpus~\cite{eslami2023pubmedclip,zhang2025MultimodalBiomedicalFoundation}.

We evaluated zero-shot performance, representational quality, and transfer capabilities across 110 tasks in six fundus imaging datasets, in particular for relatively rare diseases or pathologies for which no large-scale datasets exist, such as anterior ischemic optic neuropathy, retinitis, and chorioretinitis~\cite{pachade2021RetinalFundusMultiDisease}.
We provide the finetuned models, the human-annotated ground-truth data, and the full dataset generation pipeline as a resource for the community.

\section{Methods}\label{sec:methods}

\subsection{Study design}

We compared four sources of training data for ophthalmic vision-language models: simplistic text templates, augmented text templates, medical reports, and domain-specific scientific literature, and additionally evaluated a general biomedical literature model as a baseline.
All models we trained shared the same initialization, architecture, and training schedule, so that differences in performance can be attributed to the training data alone, while the baseline was used as released.
We set the maximum learning rate per dataset to avoid divergence. All other hyperparameters were held constant.
We evaluated the resulting models in three settings - k-NN, zero-shot, and linear probing - across 110 tasks in six fundus imaging datasets, covering both common and rare pathologies.
Furthermore, we created three variants of PubMed-Ophtha for ablation experiments.
As captions in both template datasets were generated from a fixed set of categorical labels, training on this data was effectively a classification task rather than a captioning task with diverse text.
The respective dataset sizes can be found in Table~\ref{tab:training_datasets}.

\begin{table}[htbp]
\centering
\caption{Sizes of the training datasets. For PubMed-Ophtha and its variants, the input images during training are the panels and the texts are the matching subcaptions.}
\label{tab:training_datasets}
\footnotesize
\begin{tabular}{lrrr}
\toprule
\textbf{Dataset} & \textbf{Pairs} & \textbf{Unique images} & \textbf{Unique texts} \\
\midrule
PubMed-Ophtha                  & 69{,}373  & 69{,}373  & 60{,}818 \\
-- CFP             & 13{,}243  & 13{,}243  & 12{,}197 \\
-- CFP-matched       & 9{,}427   & 9{,}427   & 8{,}850  \\
-- decontaminated & 68{,}689  & 68{,}689  & 60{,}187 \\
Simplistic templates (Kaggle EyePACS)                 & 28{,}100  & 28{,}100  & 5        \\
Augmented templates (FLAIR)                          & 406{,}757 & 210{,}321 & 212 \\
Medical reports (DeepEyeNet)                     & 22{,}162  & 9{,}428   & 3{,}148  \\
\bottomrule
\end{tabular}
\end{table}

\subsection{Domain-sepcific literature dataset construction}

The PubMed-Ophtha dataset was constructed through a four-stage pipeline (Fig.~\ref{fig:workflow}).
First, PubMed Central was filtered to identify articles relevant to retinal imaging (Fig.~\ref{fig:workflow}A, B).
Second, figures and their captions were extracted directly from the article PDFs at full resolution and enriched with additional figure-level information, such as in-text mentions, from the BIOMEDICA~\cite{lozanoBIOMEDICAOpenBiomedical} dataset (Fig.~\ref{fig:workflow}C).
Third, each figure was decomposed into its constituent panels, panel identifiers, and images, and each image was further assigned an imaging type and a mark status (Fig.~\ref{fig:workflow}D).
Fourth, figure captions were split into panel-level subcaptions and assigned to their corresponding panels (Fig.~\ref{fig:workflow}D).
The result is a hierarchically structured dataset in which each panel is associated with a subcaption and one or more images, each with an imaging type and a mark status.
Table~\ref{tab:model-overview} contains the models used in each processing step.

\begin{table}[!ht]
    \caption{Overview of the models used in the dataset processing pipeline.}
    \label{tab:model-overview}
    \footnotesize
    \begin{tabular}{
        >{\raggedright\arraybackslash}p{0.27\textwidth}
        >{\raggedright\arraybackslash}p{0.28\textwidth}
        >{\raggedright\arraybackslash}p{0.31\textwidth}
    }
    \hline
    \textbf{Step} & \textbf{Model} & \textbf{Data} \\
    \hline
    Panel Detection & RetinaNet & PubMed-Ophtha-Annotation \\
    Image Detection & RetinaNet & PubMed-Ophtha-Annotation \\
    Mark Status Classification & ResNet-50 & PubMed-Ophtha-Annotation \\
    Caption Splitting & Qwen3 32B-AWQ & PubMed-Ophtha-Annotation \\
    Optical Character Recognition & EasyOCR (CRNN~\cite{shi2017crnn}) & -- \\
    Panel Assembly & Qwen3-VL 30B-A3B & Manual \\
    \hline
    \end{tabular}
\end{table}

\subsubsection{Terminology}
\begin{figure}[!ht]
    \centering
    \includegraphics[width=0.9\textwidth]{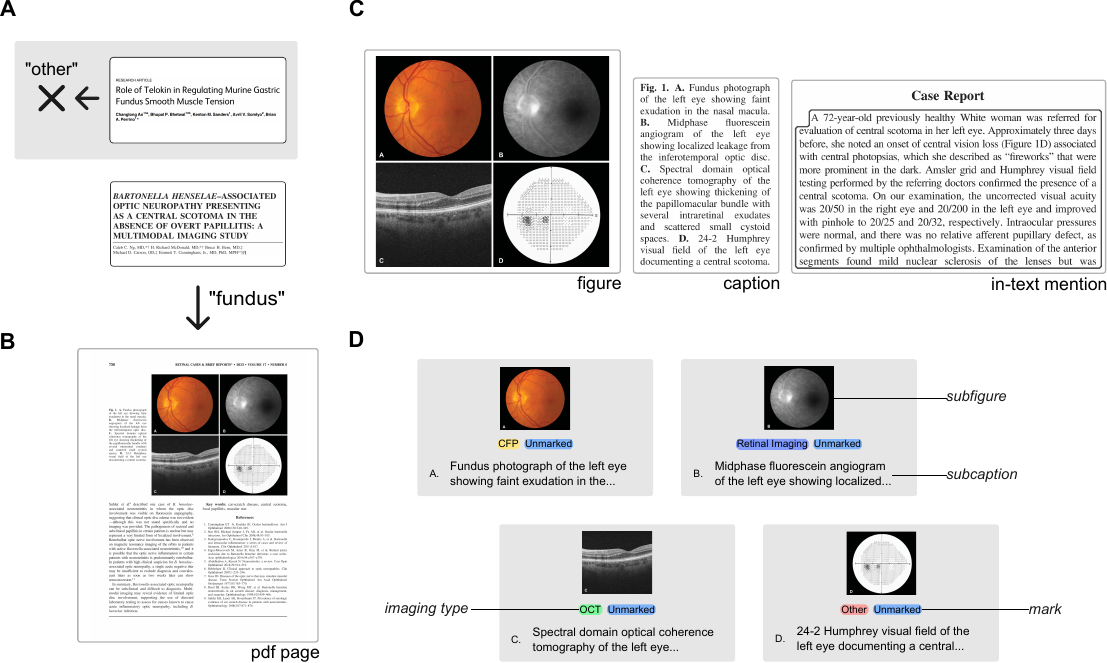}
    \caption{Overview of the dataset extraction pipeline. (A) Articles are filtered by keywords to select those relevant to ophthalmic retinal imaging. (B) A heuristic detects figures and their captions in the article PDF and extracts them at full resolution. (C) Additional figure-level information, such as in-text mentions, is retrieved from the BIOMEDICA dataset. (D) The final dataset contains individual panels with their subcaptions, imaging type annotations, and mark status.}
    \label{fig:workflow}
\end{figure}

We used the following terminology throughout (Fig.~\ref{fig:workflow}C, D).
A figure refers to the complete visual element as it appears in the publication, potentially comprising multiple components.
Each such component is called a panel and is typically distinguished by a panel identifier -- a letter (A, B, C), number, or positional descriptor (top, left, right) appearing in the caption, on the figure, or both.
Figures without any panel identifiers are treated as single-panel figures.
A panel may contain one or more images, defined here as bitmap elements within the figure, such as an ophthalmic acquisition, a graph, or a block of text.
The caption refers to the full figure caption as published; a subcaption is the portion of the caption corresponding to a single panel.
An in-text mention is any reference to a figure appearing in the article body.
Each image is assigned exactly one imaging type — \texttt{OCT} for OCT scans, \texttt{CFP} for narrow-angle color fundus photography, \texttt{Retinal Imaging} for ultra-wide field imaging and other retinal depictions such as fluorescein angiography, and \texttt{Other} for all remaining content such as graphs or ultrasound images — and a mark status indicating whether it contains a mark, defined as an arrow, dot, or similar annotation referenced in the caption or labeled directly on the figure.

\subsubsection{Filtering PubMed Central for ophthalmic articles}
PubMed Central (PMC) is one of the largest openly available corpora of biomedical literature, containing over 10 million\footnote{\url{https://pmc.ncbi.nlm.nih.gov/about/intro/}} articles across a broad range of domains.
In PMC, articles are provided either as a standalone PDF or as a package comprising the PDF, a machine-readable XML file, and the article figures as JPEG images; the latter were often compressed and therefore of lower resolution than the figures in the original PDF.
The BIOMEDICA dataset~\cite{lozanoBIOMEDICAOpenBiomedical} built on PMC by providing its figure-caption corpus as an easily processable web dataset, enriched with automatically detected figure-type labels.
We constructed our dataset by combining data from BIOMEDICA, which provided convenient access to figure captions and figure-type labels, with data extracted directly from the article PDFs available through the PMC FTP services, as well as article metadata retrieved via Entrez~\cite{2010EntrezProgrammingUtilities}.

We selected articles relevant to retinal imaging by filtering on keywords drawn from figure captions, article key terms, and Medical Subject Headings (MeSH).
The imaging keywords used were ``color fundus photography", ``CFP", ``optical coherence tomography", and ``OCT" (Fig.~\ref{fig:workflow}A).
An article was retained if it satisfied at least one criterion from each of the following three levels: (i) at least one figure or table caption contained an imaging keyword or the word ``fundus"; (ii) the article key terms included an imaging keyword or the word ``retina", or the MeSH terms included ``optical coherence tomography" or ``ophthalmoscopy"; and (iii) the full text contained ``retina" or ``ophthalmology", or the MeSH terms included ``fundus oculi" or ``retina".
Additionally, articles in the BIOMEDICA dataset containing at least one figure labeled as OCT were retained if they met criterion (ii).
After this filtering step, 24,781 articles remained from the 5 million articles with figures in the BIOMEDICA dataset.

\subsubsection{Extracting high-resolution figures from article PDFs}
To ensure high-resolution figures, we extracted figures and their captions directly from the article PDFs (Fig.~\ref{fig:workflow}B), rather than relying on the compressed JPEG images provided in the PMC article packages.
Each page was processed separately through four steps.
First, candidate figures were detected by analyzing the internal PDF layout and merging proximate bitmap and vector graphics; elements that were too small or located in the page margin were discarded as likely page header artifacts.
Second, each text segment on the page was assigned a caption likelihood score for each candidate figure.
This score increased if the text began with or contained the word ``figure", if the text was spatially close to and horizontally aligned with the figure's bounding box, and if it appeared below the figure; it decreased if other page elements intervened between the text and the figure, or if the text was positioned diagonally relative to it.
Third, figures and captions were assigned to one another based on this score and their mutual distance, with conflicts -- such as multiple captions per figure or multiple figures per caption — resolved heuristically.
Fourth, the resulting text and figure bounding boxes were refined and merged with surrounding elements where appropriate.
The figure identifier was extracted from the assigned caption using a regular expression.

Since the XML article files did not contain information about figure placement within the PDF, each extracted figure was matched to its corresponding entry in the BIOMEDICA dataset, enabling the figures to be linked back to the XML file and enriched with additional features such as in-text mentions.
Matching was based on similarity between the extracted figure and the corresponding image in the PMC package, assessed across four criteria: (i) the Hamming distance between perceptual image hashes did not exceed a threshold; (ii) the figure identifier extracted from the caption matched an identifier estimated from the package filename; (iii) the difference in aspect ratio was below a threshold; and (iv) the Levenshtein distance between the extracted caption and the caption in the BIOMEDICA dataset did not exceed a threshold.
An extracted figure was assigned to a package entry if it passed at least two of these checks; where multiple extracted figures were matched to the same entry, only the one with the greatest number of similarity matches was retained.
Figures spread across multiple pages in the PDF were joined during this step.
In total, 74,121 figures were extracted from the 18,395 articles.
The extraction process did not extract any figures from the remaining 6,386 articles.

\subsubsection{Panel and Image Detection}
\begin{figure}[!ht]
    \centering
    \includegraphics[width=0.9\textwidth]{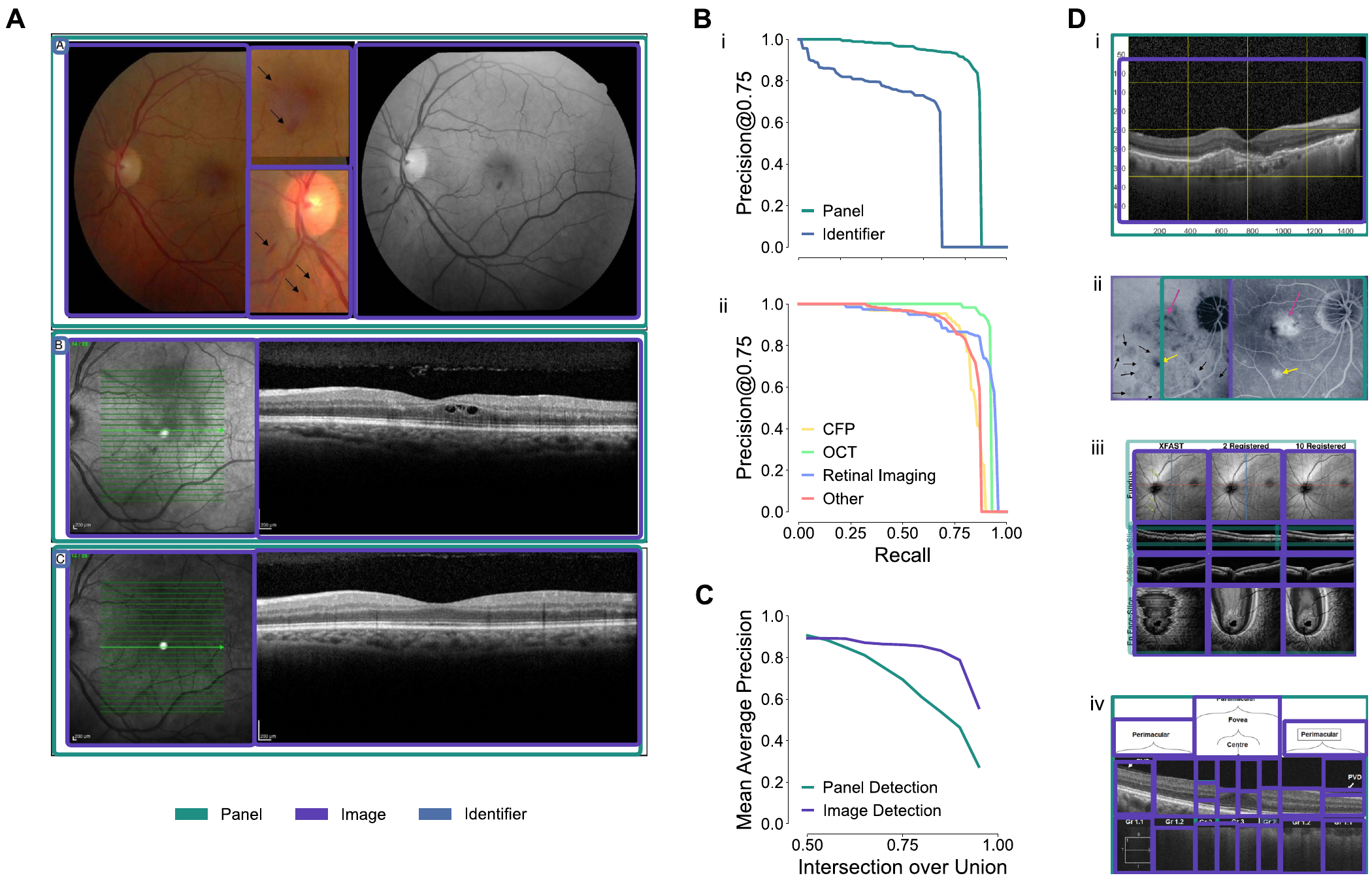}
    \caption{Detection performance and failure cases. (A) Example detections from the test set, showing panel (teal), image (purple), and panel identifier (blue) bounding boxes across three figures of varying complexity. (B) Precision-recall curves at an IoU threshold of 0.75 for (i) panel and panel identifier detection and (ii) imaging type detection across the four imaging type categories (CFP, OCT, Retinal Imaging, Other). (C) Mean average precision as a function of IoU threshold for the panel detection and image detection models. (D) Representative failure cases: (i) bounding boxes cropping the empty area at the top of OCT scans; (ii) incorrect panel division in figures lacking panel identifiers; (iii) clipping of embedded textual descriptions at panel boundaries; (iv) fragmentation of images due to confusion of visual guides with spatial separators.}
    \label{fig:detection-performance}
\end{figure}

Having filtered articles and extracted their figures, the next step was to detect panels and images within each figure, along with their imaging type and mark status, using a set of trained detection and classification models.
Existing datasets for figure parsing address only subsets of the annotations required here.
The ImageCLEF2016 dataset~\cite{de2016overview} provides bounding box annotations for individual images within figures, while the PanelSeg dataset~\cite{zou2020UnifiedDeepNeural} annotates panels and panel identifiers.
Neither dataset combines all three annotation types, and neither includes imaging type or mark status labels.
The BIOMEDICA dataset provides imaging type annotations at the panel level, but does not cover the fundus imaging modalities relevant to our setting.

We therefore collected and manually annotated a dataset of 1,901 figures using Label Studio~\cite{label_studio} in multiple annotation rounds (\hyperref[S1_Fig]{Supplementary Figs. 1} - \hyperref[S3_Fig]{3}).
In the first round, annotators were presented with each figure, its caption, and associated metadata, and were asked to correct proposed panel bounding boxes and assign imaging type and mark status labels at the panel level.
The boxes were proposed by a model trained on the ImageCLEF2016 dataset.
Where panel identifiers were positional (e.g., ``first row"), annotators used the caption to determine panel positions.
In figures with nested panel hierarchies, only the highest hierarchical level was annotated.
In the second round, annotators refined the panel annotations and added panel identifier bounding boxes.
In the final round, individual image bounding boxes were added, each assigned exactly one imaging type and one mark status, with further refinement of existing bounding boxes where necessary.
To support evaluation of the caption splitting step, the captions of 158 randomly selected figures were additionally annotated with subcaptions, textual panel identifiers, and associated panel bounding boxes.

The final annotated dataset contains panel, panel identifier, and image bounding boxes for 1,534 figures (Fig.~\ref{fig:detection-performance}A).
Of these, 158 also include per-panel subcaptions and panel identifier labels, and a further 351 contain only panel and panel identifier annotations, as the model already achieved sufficient performance on these samples.
The remaining 16 figures were excluded: these comprised figures whose panels could not be separated by bounding boxes, and figures whose subcaptions referenced marks rather than panels.
This dataset, including bounding boxes and all other annotations, is available as PubMed-Ophtha-Annotation.

Separating figures into their constituent panels and images is a multi-task problem, comprising panel detection, panel identifier detection, and image detection.
Because each image is assigned both an imaging type and a mark status, image detection is in turn a multi-label task.
To avoid the complexity of a joint multi-label detection setting -- which would prevent the use of task-specific pretraining weights and architectures, and risk conflating panel-level and image-level predictions -- we decomposed the problem into three separate models: two multi-class bounding box detectors and one binary mark status classifier.

Running two detection models independently also introduced a natural fallback, where the output of one could be used to refine the other during assignment.

For both bounding box detection tasks, we used Detectron2 ~\cite{wu2019detectron2} to fully finetune a RetinaNet~\cite{linFocalLossDense} with a ResNet-50~\cite{he2016DeepResidualLearning} backbone, pretrained on MS-COCO~\cite{lin2014MicrosoftCOCOCommon} and subsequently finetuned on the ImageCLEF2016 dataset~\cite{de2016overview}.
For each task, six models were trained with different random seeds, and the best model was selected based on validation mAP at an IoU threshold of 0.90.

The first model detected panels and panel identifiers.
We merged our annotated dataset with the PanelSeg dataset~\cite{zou2020UnifiedDeepNeural}, resampling PanelSeg to remove overlap with the ImageCLEF2016 training, validation, and test splits.
The merged dataset was used to finetune the ImageCLEF2016-pretrained model for 100,000 iterations with a batch size of 16 on 1 NVIDIA A100 GPU.
The best model reached an mAP@0.90 of 0.463 after 26,865 iterations (mAP@0.50: 0.906).
The second model detected images and assigned their imaging types.
It was finetuned from the same ImageCLEF2016-pretrained checkpoint on our collected imaging type annotations for 15,000 iterations with a batch size of 16 on 1 NVIDIA V100 GPU, reaching an mAP@0.90 of 0.787 after 3,909 iterations (mAP@0.50: 0.892).

The mark status classifier was a ResNet-50 with ImageNet~\cite{deng2009ImageNetLargescaleHierarchical} initialization, trained on our annotated dataset for 35 epochs with a batch size of 32. 
Six models were trained with different random seeds, and the best was selected based on validation accuracy, reaching 89.5\% on the test set after 7 epochs.
During training, the input image was augmented with random cropping, flips, affine transformations, and color augmentations to make the model robust to errors in the figure extraction and image detection.
The cropping augmentation operated on the full figure and randomly perturbed each corner of the ground-truth bounding box asymmetrically, shifting it outward by up to 20\% or inward by up to 30\% of the box dimensions.

During inference, the panel detection and image detection models processed each figure independently.
Detected bounding boxes were filtered using a score threshold of 0.25, followed by non-maximum suppression with an IoU threshold of 0.3; for the panel detection model, suppression was applied class-wise so that only boxes of the same type were compared.
Figures were then cropped to the detected image bounding boxes, and the mark status classifier was applied to each cropped image individually.

Across 72,802 figures, this step yielded 262,104 panels, 448,090 images, and 226,562 panel identifiers.
18 figures were rejected due to missing detections.

\subsubsection{Caption splitting}
\begin{figure}[!ht]
        \centering
    \includegraphics[width=0.9\textwidth]{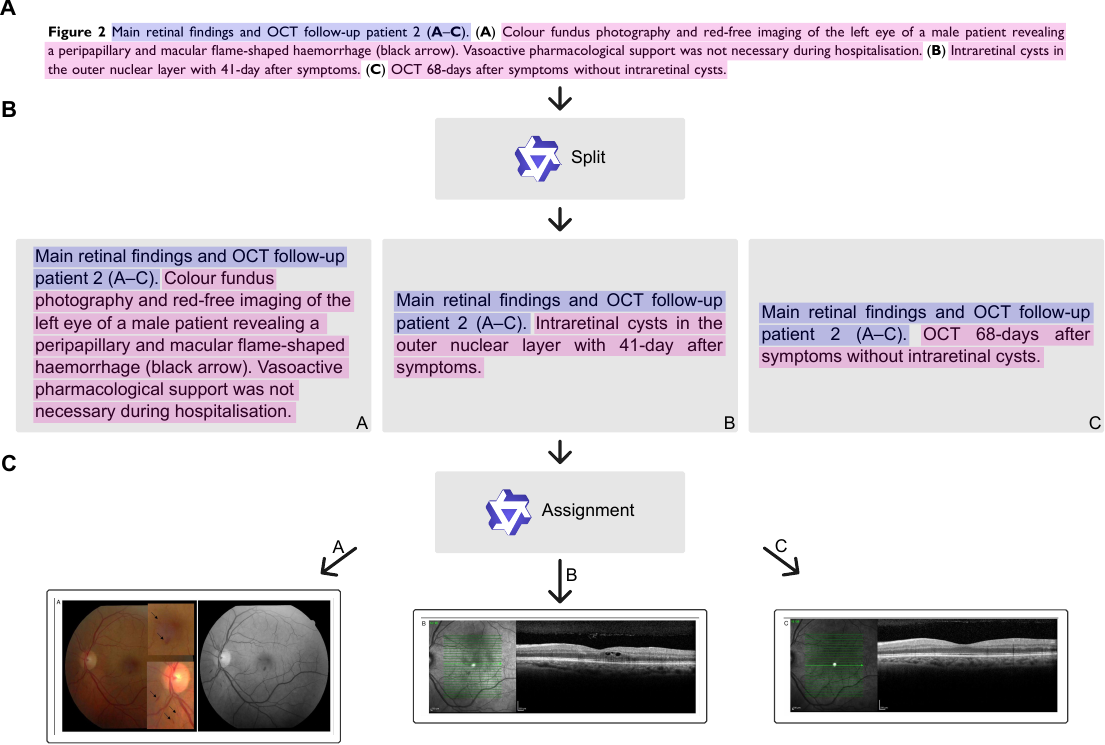}
    \caption{Caption splitting and subcaption assignment. (A) Original figure with its full caption. (B) Result of the caption splitting step: the full caption is decomposed into panel-level subcaptions, each associated with a panel identifier. (C) Result of the panel assembly step: each subcaption is assigned to its corresponding panel, and panel identifier locations are matched to the detected panel bounding boxes.}
    \label{fig:caption_split}
\end{figure}

Figure captions in medical publications are highly information-dense, often containing detailed descriptions of pathologies alongside references to specific panels.
After detecting panels within a figure, captions were split into panel-level subcaptions to preserve this semantic correspondence (Fig.~\ref{fig:caption_split}A, B).
The output of this step was a mapping from each panel identifier to its corresponding subcaption.

Given the considerable variability in caption formatting across publications, we adopted an LLM-based approach.
Initial experiments revealed that when prompted to split captions directly, the model struggled to reliably identify panel identifiers, often describing subcaption content instead.
To address this, we decomposed the task into two sequential steps: the model first extracted all panel identifiers from the full caption, and then assigned the corresponding subcaption to each identifier.
Captions for which no panel identifiers were found were treated as single-panel figures and assigned in full. 

The model was deployed using vLLM~\cite{kwon2023efficient} with Activation-aware Weight Quantization~\cite{lin2023awq} and parsed all model outputs with Pydantic~\cite{colvin2026pydantic}, a library for data validation that allows parsing of typed and structured JSON text.
Separate system prompts were used for each of the two steps\footnote{\url{https://github.com/berenslab/pubmed-ophtha/blob/main/src/pubmed_ophtha/caption_splitting/messages.py}}.
The model was provided with 17 manually annotated in-context examples for the panel identifier extraction step and 7 examples for the subcaption extraction step.
In total, 72,756 captions were split into 245,693 subcaptions, 64 captions were discarded due to parsing errors.

\subsubsection{Panel assembly}
After the previous processing steps, each figure was associated with a set of detected panels, a set of subcaptions with their panel identifiers, a set of panel identifier bounding box locations, and a set of detected images.
The goal of this step was to merge these into a unified per-panel record, where each panel is associated with a subcaption and, where applicable, one or more images and panel identifier locations.
Caption assignment was performed only for figures in which at least one image had an imaging type other than \texttt{Other}, leading to the rejection of 37,915 figures.

Assignment began with a geometric matching step.
Images and panel identifier locations were assigned to panels based on the intersection over union (IoU) and overlap fraction, using thresholds of 0.3 and 0.7, respectively.
The IoU quantifies the overlap of two boxes as the ratio of their intersection area to their union area.
For figures with alphabetic panel identifiers, subcaptions were assigned automatically by reading the panel identifier text at the detected panel identifier location using EasyOCR\footnote{\url{https://github.com/JaidedAI/EasyOCR}}.
Where no panel identifier was detected within a panel, OCR was applied to the full panel as a fallback.

Panels with missing or incomplete subcaption assignments were subsequently refined using an LLM-based approach.
We deployed the thinking version of Qwen3-VL 30B-A3B~\cite{bai2025Qwen3VLTechnicalReport} with vLLM and prompted the model independently twice per figure to increase assignment reliability.
The model inputs, prompts, and output schemas varied during LLM refinement depending on whether the panel identifiers in the caption were positional (e.g., \textit{top}, \textit{left}), allowing for more targeted processing in those cases\footnote{\url{https://github.com/berenslab/pubmed-ophtha/blob/main/src/pubmed_ophtha/panel_assembly/messages.py}}.
When the two model responses conflicted, an adjudication step was performed in which both responses were provided to the model alongside a resolution prompt.
All outputs were parsed with Pydantic, and the assignment was validated and re-prompted up to five times to ensure consistency.

In total, 98,639 panels were fully assigned, of which 67,175 were resolved automatically, 31,291 were refined by the LLM, and 173 required manual resolution.

Figures from PubMed-Ophtha-Annotation were additionally incorporated into the dataset, with ground-truth bounding boxes used in place of model predictions where available.
For these figures, the automatic detection pipeline was applied only where annotations were absent, and panels were subsequently filtered to retain only those containing at least one retinal image.
Captions were then split, and panel assembly was performed following the same procedure as for the remainder of the dataset.
Of the resulting panels, 584 originated directly from fully annotated figures; the remaining panels were assembled using a combination of ground-truth bounding boxes and automatic processing, with 1,473 assigned automatically, 1,129 refined using the LLM, and 198 resolved manually.
The final dataset consists of 102,023 panels from 35,544 figures across 15,842 articles.

\subsection{Training data sources}

\paragraph{Simplistic templates.} We used the diabetic retinopathy (DR) Kaggle dataset (Kaggle EyePACS)\footnote{\url{https://kaggle.com/competitions/diabetic-retinopathy-detection}} to create a simplistic template dataset similar to the approach used to generate the texts in CLIP-DR~\cite{CLIPDRTextualKnowledgeGuided}.
The target text was created using the template \texttt{`A fundus image showing \{severity\} diabetic retinopathy'} where the severity could be one of \texttt{`no'}, \texttt{`mild'}, \texttt{`moderate'}, \texttt{`severe'} and \texttt{`proliferative'} (Fig.~\ref{fig:dataset-types-examples}A).
The official template dataset training data was partitioned 80/20, yielding 28,100 image-text pairs for training.

\paragraph{Augmented templates.} We used the dataset augmentation method described in FLAIR~\cite{silva2025foundation} to create an augmented template dataset (Fig.~\ref{fig:dataset-types-examples}A).
From the training set, we excluded all test datasets as well as DeepEyeNet~\cite{huang2021DeepOphtMedicalReport}, the medical report dataset used in this study, to keep the training and evaluation data sources separate.
Fundus10K~\cite{wei2018laser} could not be used at all due to lack of availability.
In total, the dataset contained 229,672 image-text pairs in the training dataset, which increased to effectively 406,757 image-text pairs when factoring in the dataset augmentation method.

\paragraph{Medical reports.} We used DeepEyeNet~\cite{huang2021DeepOphtMedicalReport} as medical report dataset.
The dataset consists of fundus images and fluorescein angiography images with clinical descriptions and keywords, which were both used as targets (Fig.~\ref{fig:dataset-types-examples}B).
The medical report dataset contained 22,162 image-text pairs from the official training split created from the medical reports and the associated keywords.

\paragraph{Domain-specific literature.} We used PubMed-Ophtha as a domain-specific literature dataset.
Samples were split at the article level into training and held-out portions, so that panels from the same figure and images from the same imaging series do not appear across splits.
All experiments reported use the training split, which contained 69,373 image-text pairs (approximately 68\% of panels) from ophthalmic literature, where the image is the panel and the text the matching subcaption (Fig.~\ref{fig:dataset-types-examples}C).

\paragraph{General biomedical literature.} As a baseline, we used the trained BiomedCLIP~\cite{zhang2025MultimodalBiomedicalFoundation} model, which was trained on PMC-15M, a generalist medical literature dataset consisting of 15 million image-text pairs from PMC.
Only a small fraction of PMC is relevant to ophthalmology: of the roughly 5 million articles with figures in BIOMEDICA, 24,781 passed our filter.

\paragraph{Dataset variants.} We derived three variants of PubMed-Ophtha for the ablation experiments.
PubMed-Ophtha-CFP contains only fundus images and their associated captions, omitting other retinal imaging modalities such as OCT, and consists of 13,243 image-text pairs.
We sampled PubMed-Ophtha-CFP to approximately the same number of unique images as the medical report dataset to eliminate dataset size as a cause for performance difference, yielding 9,427 image-text pairs (PubMed-Ophtha-CFP-matched).
Finally, we searched the articles in PubMed-Ophtha for mentions of keywords and citations related to the evaluation datasets and excluded articles that matched these criteria to create PubMed-Ophtha-decontaminated (68,689 image-text pairs).
The search ensures that no images from the evaluation datasets are included in this dataset variant but potentially excludes more images than necessary.

\subsection{Vision-language model training}\label{subsec:model-training}
We trained all models using the OpenCLIP implementation\footnote{\url{https://github.com/mlfoundations/open_clip}} on one NVIDIA A100 with 40~GB of VRAM.
Each model had a ViT-B transformer~\cite{dosovitskiy2021an} as vision encoder with a patch size of 16 (ViT-B/16) and a 12-layer decoder-only transformer text encoder.
Models were initialized with the OpenAI CLIP weights~\cite{radford2021LearningTransferableVisuala}.
We trained each model for 50 epochs with a batch size of 2048 using a cosine learning rate schedule with a linear warm-up of 5 epochs and AdamW~\cite{loshchilov2018decoupled} as optimizer.
We manually set the maximum learning rate of each model.
We trained each model on three seeds and averaged the performance across the seeds.
We tested whether our findings extend to other vision encoder architectures and sizes and therefore trained models using ResNet-50~\cite{he2016DeepResidualLearning}, ResNet-50 with 4 times scaling~\cite{radford2021LearningTransferableVisuala}, ViT-B/32 and ViT-L/14~\cite{dosovitskiy2021an} as vision backbones, following the same training procedure.

\subsection{Evaluation protocol}

\begin{table}[!ht]
\centering
\caption{Construction of the evaluation prompts using templates. The text between \{\} indicates the text that was filled into the templates.
The type of task is used to select the correct template.}
\label{tab:prompt-templates-eval}
\footnotesize
\begin{tabular}{l l}
\hline
\textbf{Task type} & \textbf{Selected template} \\
\hline
Disease detection & \texttt{A fundus image showing \{-/no\} \{disease\}.} \\
Disease staging & \texttt{A fundus image showing \{disease level\}.} \\
Anomaly detection & \texttt{A fundus image showing \{a/an\} \{normal/abnormal\} \{optic disc/macula/vessels\}.} \\
Disease risk detection & \texttt{A fundus image showing a \{diseased/normal\} eye.} \\
\hline
\end{tabular}
\end{table}

\paragraph{Evaluation Datasets.} 

We evaluated the models on six fundus image datasets.
We used BRSET~\cite{nakayama2024BRSETBrazilianMultilabel}, the REFUGE2 dataset~\cite{fang2022refuge2}, APTOS 2019\footnote{\url{https://www.kaggle.com/competitions/aptos2019-blindness-detection}}, and the Eye Disease Diagnosis and Fundus Synthesis (EDDFS) dataset~\cite{xia2022eye,xia2024benchmarking} to evaluate the model's performance on more common pathologies such as age-related macular degeneration (AMD), diabetic retinopathy (DR), and glaucoma.
We selected the Retinal Fundus Multi-Disease Image Dataset version 1 (RFMiD v1)~\cite{pachade2021RetinalFundusMultiDisease} and 2 (RFMiD v2)~\cite{panchal2023RetinalFundusMultiDisease} to evaluate the performance on rare pathologies.
We defined a task as a unique combination of an evaluation dataset and a disease or pathology.
In total, we compared the models on 110 tasks.
The respective dataset splits are detailed in \nameref{S1_Table}.

\paragraph{Metrics.} The \textit{k-NN performance} reflects how informative the resulting embedding spaces of the models are.
For this purpose, we trained a k-NN classifier on the frozen training set image embeddings of the models.
The classifiers used a distance-weighted vote, and $k$ was selected for each evaluation dataset.
We computed the \textit{zero-shot performance} for the trained models by creating template sentences (Table~\ref{tab:prompt-templates-eval}) for the presence and absence of each disease.
For each disease, we selected the option with the highest cosine similarity of these templates with the image.
For non-binary labels, templates for each class were created.
This setting is the only one that involves the text encoder, as it requires the image and text embeddings to be aligned for the evaluated labels.
As a consequence, either models trained on captions that resemble the evaluation prompts or models trained on more diverse text are favored here.
We also evaluated the \textit{linear probing performance} of the models by training a balanced linear classifier on the image embeddings.
For each dataset-model combination, the $l_2$-regularization strength was selected for maximum balanced accuracy on the validation set.
The results of this evaluation reflect the linear separability of the resulting embedding spaces.
For all metrics, we reported AUROC on the test set to mitigate class imbalance, as some diseases (e.g., chorioretinitis and asteroid hyalosis) were rare in the datasets. 
We report the mean AUROC across all tasks as well as the mean AUROC within select task groups: each task group comprises the tasks that evaluate the same condition across different evaluation datasets, independently of the task type.
These groups are namely disease risk, DR, AMD, glaucoma, myopia, hypertensive retinopathy (HR), asteroid hyalosis (AH), and macular scar (MS).
AH and MS were selected for their few positive samples in the test datasets (6 and 14 out of 810, respectively).
A full list of task groups and the tasks they comprise can be found in \nameref{S2_Table}.
In addition to the AUROC, we used a head-to-head win rate of the models trained with PubMed-Ophtha compared to others.
When comparing two models, for each task, the model with the higher AUROC is awarded a win; the win rate is the percentage of won tasks for each model.

\paragraph{Statistical analysis.} We performed a paired Wilcoxon test across all 110 tasks using Holm correction for multiple comparisons for the performance evaluation. 
We performed a two-sided binomial test with Holm correction for each win rate comparison.

\section{Results}

\subsection{Dataset composition and validation}

\subsubsection{Data overview}
The final dataset contains 102,023 panels from 35,544 figures across 15,842 open-access articles (\nameref{S4_Fig}).
Key statistics on imaging type distribution, mark status, subcaption length, and panel assembly method are summarized in Table~\ref{tab:data-overview}.

\begin{table}[!ht]
    \centering
    \caption{Overview of the PubMed-Ophtha dataset statistics.}
    \label{tab:data-overview}
    \footnotesize
    \begin{tabular}{
        >{\raggedright\arraybackslash}p{0.55\textwidth}
        >{\raggedright\arraybackslash}p{0.32\textwidth}
    }
    \hline
    \textbf{Property} & \textbf{Value} \\
    \hline
    \multicolumn{2}{l}{\textit{Dataset size}} \\
    Articles & 15,842 \\
    Figures & 35,544 \\
    Panels & 102,023\\
    Number of unique MeSH terms & 5,402 \\
    \hline
    \multicolumn{2}{l}{\textit{Imaging type distribution}} \\
    CFP & 24,155 (23.7\% of panels) \\
    OCT & 43,352 (42.5\% of panels) \\
    Retinal Imaging & 48,974 (48.0\% of panels) \\
    Other & 10,705 (10.5\% of panels) \\
    \hline
    \multicolumn{2}{l}{\textit{Mark status}} \\
    Panels with marks & 27,534 (27.0\% of panels) \\
    \hline
    \multicolumn{2}{l}{\textit{Subcaptions}} \\
    Mean length (words) & 34.1 \\
    Median length (words) & 28.0 \\
    \hline
    \multicolumn{2}{l}{\textit{Panel assembly method}} \\
    Automatic & 68,648 (67.3\%) \\
    LLM-refined & 32,420 (31.8\%) \\
    Manual & 371 (0.4\%) \\
    PubMed-Ophtha-Annotation & 584 (0.6\%) \\
    \hline
    \end{tabular}
\end{table}

\subsubsection{Technical validation}
We validated every step of the dataset extraction pipeline carefully (Fig.~\ref{fig:workflow}), treating the performance of each trained model on held-out human-annotated data as a proxy for the resulting dataset quality.
Validation covered figure extraction, panel and image detection, imaging type and mark status classification, caption splitting, and caption assignment.

Both detection models performed well overall, reliably localizing panels, images, and panel identifiers across a range of figure layouts (Fig.~\ref{fig:detection-performance}A).
We evaluated performance on a held-out test set of 188 figures (panel detection), 157 figures (image detection), and 1,054 images (mark status classification) using precision-recall curves at an IoU threshold of 0.75 and mean average precision (mAP) as a function of IoU (Fig.~\ref{fig:detection-performance}B, C).
Panel and image detection were both robust across IoU thresholds, with mAP for the panel class degrading from 0.909 to 0.532 and from 0.892 to 0.558 between IoU 0.50 and 0.95, respectively, and precision remaining high across most of the recall range (Fig.~\ref{fig:detection-performance}Bi, C).
Panel identifier detection was considerably weaker, with mAP dropping from 0.903 at IoU 0.50 to 0.018 at IoU 0.95 and precision deteriorating sharply beyond a recall of 0.6 (Fig.~\ref{fig:detection-performance}Bi).
This disparity indicates that panel identifier detection is the primary source of overall performance degradation, although errors at this stage can be partially compensated during panel assembly.
Precision-recall curves were stable across the CFP, OCT, and Retinal Imaging categories, while the \texttt{Other} category showed lower performance across all IoU values (mAP@0.50: 0.855), likely reflecting its greater visual heterogeneity (Fig.~\ref{fig:detection-performance}Bii).
The mark status classification model reached an accuracy of $89.5\%$ on the test set.

The models handled most figure layouts well, as reflected in the detection scores above.
Inspecting the remaining failure cases qualitatively, we found recurring patterns that can be partially addressed downstream (Fig.~\ref{fig:detection-performance}D).
In OCT images, bounding boxes can crop the empty area at the top of the scan, though the diagnostically relevant content is mostly retained (Fig.~\ref{fig:detection-performance}Di).
In figures lacking panel identifiers, panels are prone to fragmentation or incorrect division, but these cases are amenable to correction through LLM-based refinement during panel assembly (Fig.~\ref{fig:detection-performance}Dii).
In complex layouts, embedded textual descriptions can be clipped at panel boundaries, and marks or visual guides may be detected as separate \texttt{Other}-type images or fragment image bounding boxes across spatial separators (Fig.~\ref{fig:detection-performance}Diii, Div).
We manually annotated the ground-truth locations of the figures in the collected dataset and measured a median IoU of 0.997.

Caption splitting involves two sequential tasks (Fig.~\ref{fig:caption_split}): extracting panel identifiers from the full caption, and assigning the corresponding subcaption to each identifier.
We evaluated both tasks on the human-annotated subset of the dataset.
During evaluation, predicted subcaptions were matched to ground-truth values by normalizing and aligning panel identifiers; only samples where all panel identifiers could be matched were used to assess subcaption quality.
The proportion of unprocessed samples serves as a proxy for panel identifier extraction performance.

The two-step decomposition substantially improved panel identifier extraction.
In the few-shot setting with three examples, extracting panel identifiers as a separate first step reduced the proportion of unprocessed samples from 18.4\% to 10.6\%.
Increasing the number of few-shot examples for the identifier extraction step from three to thirteen reduced this further to 7.8\%.
The remaining errors fell into three categories: missing panel identifiers, extraction of additional hierarchical levels, and identifier mismatches requiring more complex normalization.
The latter two can be resolved during panel assembly, and samples affected by the first can often be at least partially processed.

Subcaption quality was measured using the mean average sentence BLEU score (maB), computed per sample using SacreBLEU~\cite{post2018acall} and the sentence BLEU metric~\cite{papineni2002BleuMethodAutomatic}.
Performance improved markedly with the addition of few-shot examples: enabling thinking mode in the zero-shot setting raised the maB from 0.634 to 0.682, and providing three examples increased it to 0.887.
Introducing the two-step process had only a minimal effect on the maB itself (0.885 at 3-shot, 0.876 at 13-shot), likely because the gain in processable samples partially offset the per-sample score.
The final configuration -- 17 few-shot examples for panel identifier extraction and 7 for subcaption assignment -- achieved an maB of 0.913 with 6.4\% of samples unprocessed, demonstrating that high subcaption quality and broad coverage can be achieved simultaneously.
Common errors were the failure to remove the panel identifiers from the subcaption text, failure to add text that concerns all panels to the subcaptions, such as introductory phrases or lists of abbreviations, and slight rephrasing.
As an example of this rephrasing, the model extracts the subcaption ``SEM image of sample iPP/CuNPs 0.25 wt \% at 5000× magnification." from ``SEM images with different magnification (a) 5000, (b) 30,000, (c) 30,000, and (d) 100,000 times of sample iPP/CuNPs 0.25 wt \%."~\cite{jardon2021antimicrobial}.
The corresponding ground-truth annotation ``SEM images with magnification 5000 times of sample iPP/CuNPs 0.25 wt \%." is semantically equivalent to the predicted subcaption.
Overall, these errors have only a minor impact on dataset quality, as the resulting subcaptions remain semantically equivalent to the ground truth in the majority of cases.

The LLM refinement adds a fallback layer for errors in the extracted panel identifiers, as slight identifier mismatches and superfluous subcaptions can be resolved here.
Prompting the model twice and an optional third adjudication step increase the stability of the assignment output.
Furthermore, the functionality of the caption assignment was validated on a randomly sampled selection of 50 figures, out of which 25 were automatically assigned and 25 used LLM refinement.
In total, 88\% of the panels were correctly assigned, and 6\% were incorrectly assigned.
The remaining 6\% of the entries had panel bounding boxes that spanned multiple panels and could therefore not be evaluated.
In those cases, the assigned panel was among the spanned panels.
Notably, all of the automatically assigned panels in the test had the correct assignment.

\subsection{Effect of training data source on model performance}

Having established that PubMed-Ophtha captures ophthalmic panels and their subcaptions with high fidelity, we next evaluated whether this translates into measurable gains in transferability and representational quality over models trained on the other training data sources (Fig.~\ref{fig:dataset-types-examples}).
The compared sources were domain-specific literature (PubMed-Ophtha), simplistic templates, augmented templates, medical reports, and general biomedical literature.
To this end, we evaluated their zero-shot, k-NN, and linear probing AUROC in six challenging fundus imaging datasets with a total of 110 clinical tasks, including detection of ophthalmic diseases and other pathologies of the retina (see \nameref{sec:methods}).

\begin{figure}
    \centering
    \includegraphics[width=\linewidth]{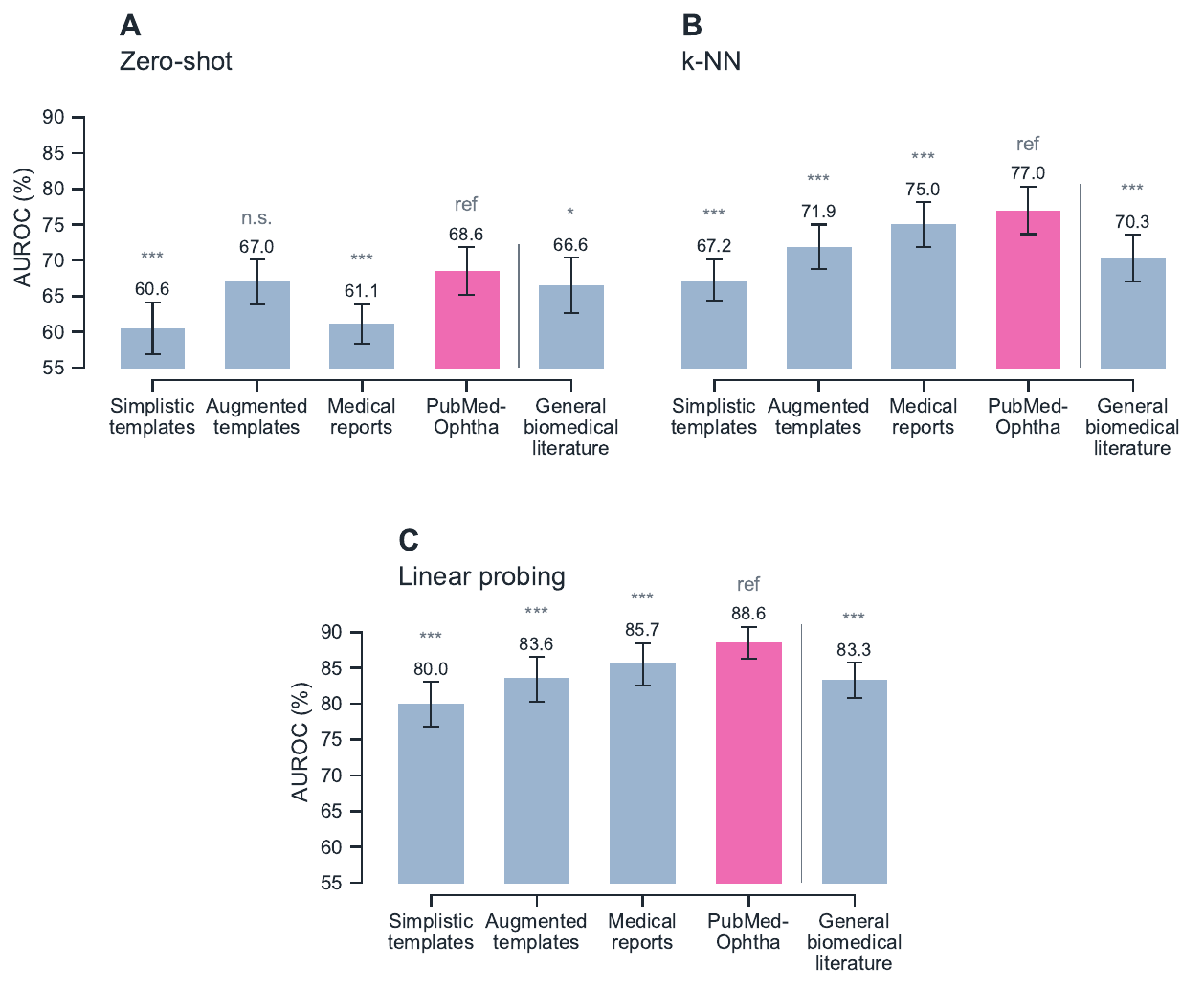}
    \caption{Performance comparison of training data sources across tasks and evaluation methods. Each bar shows the average AUROC across tasks with 95\% confidence interval.
    The stars above the plots indicate the significance level in relation to PubMed-Ophtha ($^{*}p<0.05$, $^{**}p<0.01$, $^{***}p<0.001$; ns = not significant ($p\geq0.05$); ref = reference). (A) Zero-shot AUROC. (B) Distance-weighted k-NN AUROC on the frozen vision encoder embeddings. (C) Balanced linear probing AUROC on the frozen vision encoder embeddings.}
    \label{fig:performance-overview}
\end{figure}

\begin{figure}
    \centering
    \includegraphics[width=\linewidth]{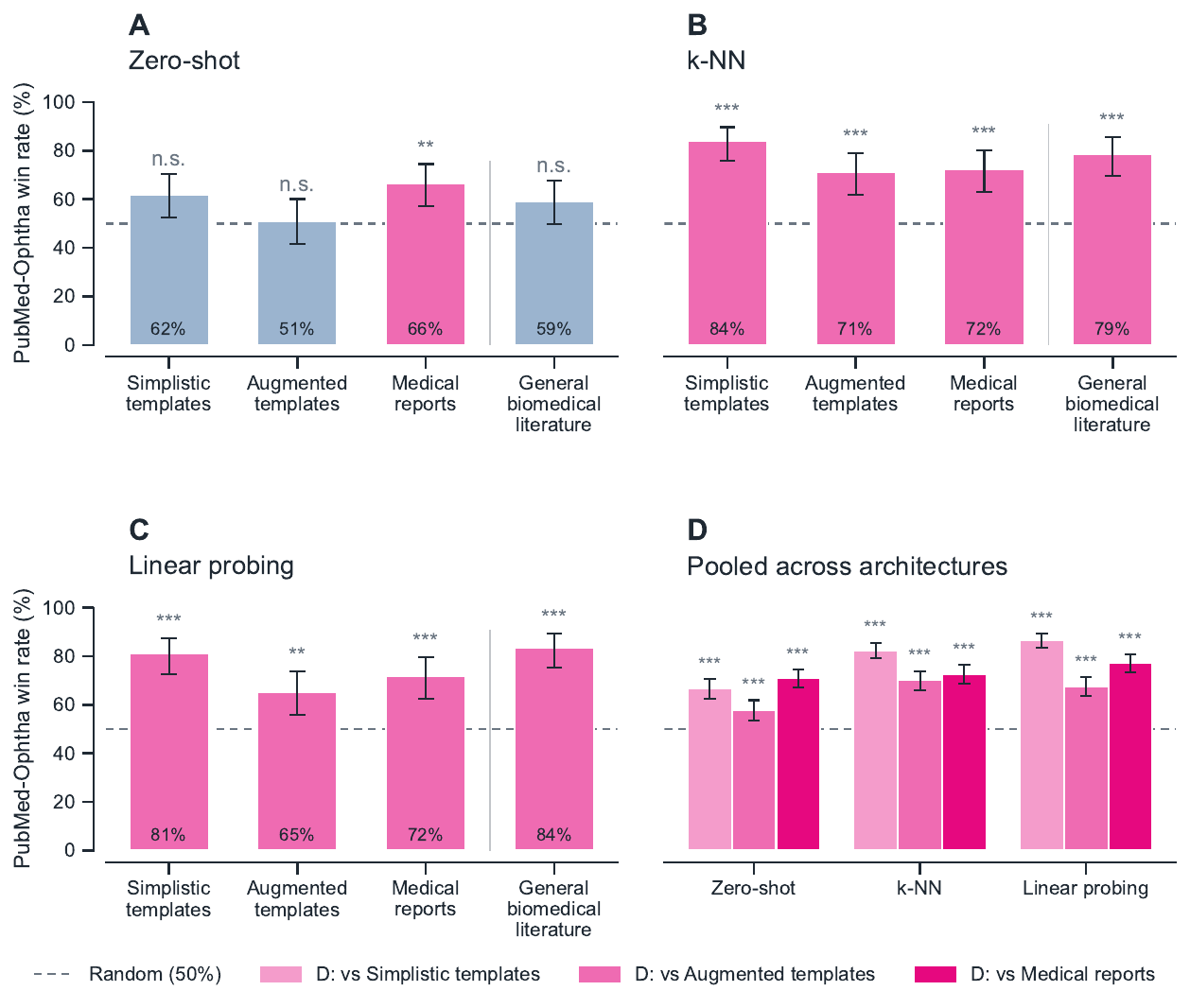}
    \caption{Win rate of models trained on PubMed-Ophtha against to others. Each bar shows the percentage of tasks won by PubMed-Ophtha.
    Wilson 95\% confidence intervals.
    The stars above the plots indicate the significance level ($^{*}p<0.05$, $^{**}p<0.01$, $^{***}p<0.001$; ns = not significant ($p\geq0.05$)). (A) Win rate in the zero-shot evaluation when using ViT-B/16 as vision encoder. (B) Win rate of the distance-weighted k-NN classifier when using ViT-B/16 as vision encoder. (C) Win rate in the linear probing evaluation when using ViT-B/16 as vision encoder. (D) Win rates pooled over all vision encoder architectures (ResNet-50, ResNet-50x4, ViT-B/16, ViT-B/32, ViT-L/14). }
    \label{fig:performance-winrate}
\end{figure}

\begin{table}[tb]
\centering
\caption{Zero-shot scores on select task groups. Each score is the mean AUROC (\%, $\uparrow$) across all tasks that fall into the task group.
The average column indicates the mean score across all 110 tasks.
Bold indicates the best, and underlined the second-best scores among comparable training data sources.
AMD: age-related macular degeneration, HR: hypertensive retinopathy, AH: asteroid hyalosis, MS: macular scars. $\ddagger$The augmented template dataset directly overlaps the classification labels for AMD, glaucoma, myopia, HR, AH, and MS specifically, so those scores reflect in-domain classification rather than generalization to unseen labels.}
\label{tab:auroc_disease_zero_shot}
\footnotesize
\setlength{\tabcolsep}{4pt}
\resizebox{\ifdim\width>\linewidth\linewidth\else\width\fi}{!}{
\begin{tabular}{lcccccccc}
\toprule
\textbf{Dataset/Model} & \textbf{Disease risk} & \textbf{AMD} & \textbf{Glaucoma} & \textbf{Myopia} & \textbf{HR} & \textbf{AH} & \textbf{MS} & \textbf{Average} \\
\midrule
Simplistic templates & \underline{77.52} & \underline{55.16} & \underline{53.97} & 47.07 & \textbf{60.73} & \underline{65.21} & 47.52 & 60.56 \\
Augmented templates & 76.49 & 64.09$^\ddagger$ & 70.44$^\ddagger$ & 64.39$^\ddagger$ & 69.45$^\ddagger$ & 86.92$^\ddagger$ & 61.62$^\ddagger$ & \underline{67.05} \\
Medical reports & 75.64 & 53.19 & 50.46 & \underline{59.60} & \underline{59.23} & 49.66 & \underline{50.62} & 61.13 \\
PubMed-Ophtha & \textbf{79.64} & \textbf{58.19} & \textbf{61.27} & \textbf{78.90} & 46.46 & \textbf{84.85} & \textbf{76.43} & \textbf{68.55} \\
\midrule
General biomedical literature (baseline) & 78.81 & 72.63 & 56.38 & 87.75 & 49.48 & 90.42 & 69.35 & 66.58 \\
\bottomrule
\end{tabular}
}
\end{table}

\paragraph{Zero-shot performance.} We found that training with domain-specific literature datasets on average led to significantly better zero-shot AUROC than general biomedical literature, simplistic templates and medical reports, while the augmented templates did not result in significantly different performance in the zero-shot setting (Fig.~\ref{fig:performance-overview}B)\footnote{Statistical results: PubMed-Ophtha vs. Simplistic templates: $p<0.001$; PubMed-Ophtha vs. Augmented templates: $p=0.147$; PubMed-Ophtha vs. Medical reports: $p<0.001$; PubMed-Ophtha vs. General literature: $p=0.037$; $n=110$, paired Wilcoxon test}.
In general, zero-shot performance varied between tasks and training data sources.
The simplistic template and medical report datasets dropped to near chance-level performance on most task groups, with an average zero-shot performance slightly higher than 60\% across all 110 clinical tasks (Table~\ref{tab:auroc_disease_zero_shot}).
In contrast, both literature approaches performed well, even on rare conditions, and had a total average of 68.55\% and 66.58\% for PubMed-Ophtha and the general biomedical literature baseline, respectively.
Models trained on augmented templates performed well with an average AUROC of 67.05\%, but were often outperformed by other models on the AMD, myopia, AH, and MS task groups, even though evaluation collapses to classification on those task groups.
Notably, PubMed-Ophtha performed worst on the HR task group in the zero-shot setting, with an AUROC of 46.46\% and thus below chance level, even though it reached the best HR score in the k-NN evaluation (Table~\ref{tab:auroc_disease_knn}).
In the zero-shot evaluation on the DR task group, PubMed-Ophtha outperformed augmented templates, despite this model being mainly trained on DR disease staging labels (Table~\ref{tab:auroc_dr_by_setting}). Also, the model using domain-specific literature won significantly more tasks than models based on medical reports (Fig.~\ref{fig:performance-winrate}B).
While PubMed-Ophtha had a higher win rate on average, only the comparison against medical reports was significant\footnote{vs. Simplistic templates: $p=0.05$; vs. Augmented templates: $p=0.924$; vs. Medical reports: $p = 0.01$; vs. General literature: $p=0.139$; $n=110$, two-sided binomial test}.

\begin{table}[tb]
\centering
\caption{k-NN scores on select task groups. Each score is the mean AUROC (\%, $\uparrow$) across all tasks that fall into the task group.
The average column indicates the mean score across all 110 tasks.
Bold indicates the best, and underlined the second-best scores among comparable training data sources.
AMD: age-related macular degeneration, HR: hypertensive retinopathy, AH: asteroid hyalosis, MS: macular scars. $\ddagger$The augmented template dataset directly overlaps the classification labels for AMD, glaucoma, myopia, HR, AH, and MS specifically, so those scores reflect in-domain classification rather than generalization to unseen labels.}
\label{tab:auroc_disease_knn}
\footnotesize
\setlength{\tabcolsep}{4pt}
\resizebox{\ifdim\width>\linewidth\linewidth\else\width\fi}{!}{%
\begin{tabular}{lcccccccc}
\toprule
\textbf{Training data source} & \textbf{Disease risk} & \textbf{AMD} & \textbf{Glaucoma} & \textbf{Myopia} & \textbf{HR} & \textbf{AH} & \textbf{MS} & \textbf{Average} \\
\midrule
Simplistic templates & 90.49 & 63.47 & 63.59 & 86.40 & \underline{71.95} & 57.99 & 55.34 & 67.24 \\
Augmented templates & 92.12 & 75.44$^\ddagger$ & 78.09$^\ddagger$ & 86.07$^\ddagger$ & 73.41$^\ddagger$ & 80.97$^\ddagger$ & 70.27$^\ddagger$ & 71.85 \\
Medical reports & \underline{92.34} & \textbf{84.58} & \underline{75.78} & \underline{95.86} & 69.25 & \underline{87.72} & \underline{67.52} & \underline{75.03} \\
PubMed-Ophtha & \textbf{93.38} & \underline{79.45} & \textbf{76.56} & \textbf{96.52} & \textbf{76.36} & \textbf{93.07} & \textbf{77.14} & \textbf{77.02} \\
\midrule
General biomedical literature (baseline) & 90.64 & 74.37 & 68.30 & 93.20 & 64.16 & 89.31 & 59.01 & 70.34 \\
\bottomrule
\end{tabular}
}
\end{table}

\paragraph{k-NN performance.} We found that training on PubMed-Ophtha resulted in significantly better k-NN scores compared to all other training data sources (Fig.~\ref{fig:performance-overview}A)\footnote{Statistical results: PubMed-Ophtha vs. other training data sources $p<0.001$ for each comparison; $n=110$, paired Wilcoxon test}.
When inspecting the task groups, we found that PubMed-Ophtha performed best on most of the groups, only placing second-best in the AMD task group (Table~\ref{tab:auroc_disease_knn}).
The training labels of the augmented templates directly overlap the labels evaluated in all but the disease risk task group.
In those cases, the scores reflect in-domain classification rather than transfer to unseen labels.
Notably, PubMed-Ophtha still outperformed the augmented templates in most of these groups.
On average, PubMed-Ophtha reached a k-NN AUROC of 77.02\% across all 110 clinical tasks, ahead of medical reports (75.03\%), augmented templates (71.85\%), general biomedical literature (70.34\%), and simplistic templates (67.24\%).
On the DR task group, PubMed-Ophtha performed almost on par with the template sources, which were mainly trained on DR disease staging labels (Table~\ref{tab:auroc_dr_by_setting}).
The win rate evaluation shows that PubMed-Ophtha wins significantly more tasks than the competing models, with a win rate of at least 71\% (Fig.~\ref{fig:performance-winrate}A).

\begin{table}[tb]
\centering
\caption{Linear probing scores on select task groups. Each score is the mean AUROC (\%, $\uparrow$) across all tasks that fall into the task group.
The average column indicates the mean score across all 110 tasks.
Bold indicates the best, and underlined the second-best scores among comparable training data sources.
AMD: age-related macular degeneration, HR: hypertensive retinopathy, AH: asteroid hyalosis, MS: macular scars. $\ddagger$The augmented template dataset directly overlaps the classification labels for AMD, glaucoma, myopia, HR, AH, and MS specifically, so those scores reflect in-domain classification rather than generalization to unseen labels.}
\label{tab:auroc_disease_linear_probing}
\footnotesize
\setlength{\tabcolsep}{4pt}
\resizebox{\ifdim\width>\linewidth\linewidth\else\width\fi}{!}{%
\begin{tabular}{lcccccccc}
\toprule
\textbf{Dataset/Model} & \textbf{Disease risk} & \textbf{AMD} & \textbf{Glaucoma} & \textbf{Myopia} & \textbf{HR} & \textbf{AH} & \textbf{MS} & \textbf{Average} \\
\midrule
Simplistic templates & 92.47 & 82.51 & 77.09 & 92.98 & 77.48 & 91.04 & 77.57 & 80.00 \\
Augmented templates & \underline{93.33} & 89.16$^\ddagger$ & 87.38$^\ddagger$ & 94.93$^\ddagger$ & 78.74$^\ddagger$ & 92.02$^\ddagger$ & 82.05$^\ddagger$ & 83.58 \\
Medical reports & 93.32 & \textbf{94.69} & \underline{84.86} & \underline{98.66} & \underline{78.92} & \underline{97.64} & \underline{87.90} & \underline{85.68} \\
PubMed-Ophtha & \textbf{93.83} & \underline{93.81} & \textbf{86.35} & \textbf{98.76} & \textbf{82.10} & \textbf{98.16} & \textbf{88.90} & \textbf{88.63} \\
\midrule
General biomedical literature (baseline) & 90.80 & 83.96 & 78.73 & 97.85 & 71.88 & 92.91 & 82.44 & 83.32 \\
\bottomrule
\end{tabular}
}
\end{table}

\paragraph{Linear probing performance.} We found that training with PubMed-Ophtha led to significantly better linear probing AUROC than any of the other datasets (Fig.~\ref{fig:performance-overview}C)\footnote{Statistical results: PubMed-Ophtha vs. other training data sources $p<0.001$ for each comparison; $n=110$, paired Wilcoxon test}.
On average, the model trained with this dataset reached a linear probing AUROC of 88.63\% across all 110 clinical tasks, showing very good potential of the learned representation to support high-quality read-outs ($+11.61$ and $+20.08$ percentage points over the k-NN and zero-shot settings, Table~\ref{tab:auroc_disease_linear_probing}).
Overall, all other training data sources benefited from learning a linear read-out.
PubMed-Ophtha performed best across most task groups, only placing second best next to medical reports for the AMD task group.
It also outperformed augmented templates on all task groups except the glaucoma and DR task groups, where it performed similarly ($-1.03$ and $-0.88$ percentage points, respectively), which is expected given that the augmented template model was directly trained on DR disease staging and glaucoma labels.
The PubMed-Ophtha models showed a significantly higher win rate compared to all other methods, outperforming general biomedical literature in 84\% of all tasks on average (Fig.~\ref{fig:performance-winrate}C)\footnote{vs. Augmented templates: $p=0.002$; vs. remaining training data sources: $p < 0.001$ for each comparison; $n=110$, two-sided binomial test}.

\begin{table}[tb]
\centering
\caption{Evaluation scores for diabetic retinopathy. Each score is the mean AUROC (\%, $\uparrow$) across diabetic retinopathy classification tasks.
The template datasets directly overlap the diabetic retinopathy labels evaluated, so those scores reflect in-domain classification rather than generalization to unseen labels, and are not directly comparable to the other training data sources.
Bold indicates the best score among comparable training data sources.}
\label{tab:auroc_dr_by_setting}
\footnotesize
\setlength{\tabcolsep}{4pt}
\resizebox{\ifdim\width>\linewidth\linewidth\else\width\fi}{!}{%
\begin{tabular}{llccc}
\toprule
\textbf{Group} & \textbf{Dataset/Model} & \textbf{k-NN} & \textbf{Zero-shot} & \textbf{Linear probing} \\
\midrule
Label overlap & Simplistic templates & 92.01 & 85.64 & 94.99 \\
              & Augmented templates & 92.50 & 76.61 & 95.05 \\
\midrule
No label overlap & Medical reports & 87.67 & 60.37 & 92.68 \\
                 & PubMed-Ophtha & \textbf{91.44} & \textbf{76.67} & \textbf{94.17} \\
\midrule
Baseline & General biomedical literature & 85.48 & 71.29 & 88.66 \\
\bottomrule
\end{tabular}
}
\end{table}

\paragraph{Ablations.} We investigated which parts of the PubMed-Ophtha dataset were responsible for its good performance on the diverse clinical tasks.
We compared models trained on each of the PubMed-Ophtha variants to those trained on the full dataset using two one-sided Wilcoxon tests with Holm correction against an AUROC difference margin of 2 points (Table~\ref{tab:auroc_pmo_variants}).
For the k-NN and linear probing settings, all variants were equivalent to the full dataset within this margin.
In the zero-shot setting, equivalence held only for the decontaminated variant; the CFP and CFP-matched variants performed slightly better than the full dataset on average, and equivalence within the margin could not be established.
None of these restrictions reduced performance beyond the equivalence margin, indicating that the observed gains are not explained by dataset size, by overlap with the evaluation datasets, or by the additional imaging modalities.

Across the additional vision encoder architectures, the evaluation showed that models trained on PubMed-Ophtha had a significantly higher win rate (Fig.~\ref{fig:performance-winrate}D), making our findings stable across architectures\footnote{vs. all other training data sources: $p < 0.001$ for each comparison; $n=550$, two-sided binomial test}.

\begin{table}[tb]
\centering
\caption{Performance comparison between PubMed-Ophtha variants. Each score is the mean AUROC (\%, $\uparrow$) across all 110 clinical tasks.
Bold indicates the best, and underlined the second-best score per setting. $p(\pm2)$ is the TOST (two one-sided Wilcoxon tests) p-value for that variant against PubMed-Ophtha at a $\pm$2-point margin, Holm-corrected across the 3 variants within each setting; a significant $p$ is a positive claim that the difference lies within that margin.}
\label{tab:auroc_pmo_variants}
\footnotesize
\setlength{\tabcolsep}{4pt}
\resizebox{\ifdim\width>\linewidth\linewidth\else\width\fi}{!}{%
\begin{tabular}{lcccccc}
\toprule
\textbf{Dataset} & \textbf{k-NN} & $p(\pm2)$ & \textbf{Zero-shot} & $p(\pm2)$ & \textbf{Linear probing} & $p(\pm2)$ \\
\midrule
PubMed-Ophtha & 77.02 & -- & 68.55 & -- & \underline{88.63} & -- \\
\midrule
-- CFP & \textbf{77.98} & $<$0.001 & \textbf{70.66} & 0.391 & 88.12 & $<$0.001 \\
-- CFP-matched & \underline{77.52} & $<$0.001 & \underline{70.11} & 0.282 & 88.25 & $<$0.001 \\
-- decontaminated & 76.64 & $<$0.001 & 69.48 & 0.007 & \textbf{88.70} & $<$0.001 \\
\bottomrule
\end{tabular}
}
\end{table}

\section{Discussion}
In this study, we compared how the source and content of the training data shapes the representations of ophthalmic CLIP models. We found that domain-specific literature achieved the highest average performance in all three evaluation settings, with a significant gain over simplistic templates, augmented templates, medical reports, and a general biomedical literature baseline in the k-NN and linear probing evaluations.
To this end, we collected PubMed-Ophtha, a hierarchical dataset of high-resolution ophthalmic panels with their subcaptions from open-access literature.
Through extensive technical validation, we showed the quality of the automatic collection process.
The evaluation on task groups showed similar or better performance than template-based approaches on common pathologies such as AMD and DR, and improved performance on rare pathologies such as AH and MS, even though the evaluation collapses to classification for the template-based approaches.
Notably, PubMed-Ophtha achieved this with a substantially smaller dataset size, roughly one-sixth of the effective size of the augmented template dataset.

As expected from the evaluation setup, the advantage of PubMed-Ophtha was smaller in the zero-shot setting, where models trained on more diverse text or on captions resembling the evaluation prompts can leverage their text encoder more.
This applies in particular to the general biomedical literature baseline, which was not matched to the other training data sources: BiomedCLIP was trained on roughly 200 times as many image-text pairs as the PubMed-Ophtha training split and additionally uses PubMedBERT~\cite{gu2021domain}, a domain-specific text encoder.
Notably, domain-specific literature still resulted in significantly better zero-shot performance than this baseline.
As the template models only ever see captions generated from a fixed set of categorical labels, their advantage should disappear for more diverse text, such as medical reports or patient descriptions.
Training on PubMed-Ophtha improved the representational quality of the vision embeddings independently of the vision encoder architecture (Fig.~\ref{fig:performance-winrate}D).

Existing ophthalmic CLIP models have primarily modified the architecture and the loss function rather than the training data.
CLIP-DR~\cite{CLIPDRTextualKnowledgeGuided} uses a rank-aware loss function to improve the representation of diabetic retinopathy grades in the embedding space, while RET-CLIP~\cite{du2024RETCLIPRetinalImage} directly maximizes the cosine similarity between the embeddings of the images of each eye and the text.
EyeCLIP~\cite{shi2025multimodal} combines CLIPs with a masked auto-encoder, using both image-text and cross-image modality contrastive loss.
Our results indicate that the choice of the training data source should be considered alongside these modifications, as it affects representational quality independently of the architecture.

As our evaluation covers fundus images only, the OCT and other retinal imaging pairs cannot contribute directly to these tasks.
We nonetheless release and train on the full dataset, as OCT and other retinal imaging modalities are required for applications that our evaluation datasets do not cover.
Beyond the panels and their subcaptions, PubMed-Ophtha records the presence of marks such as arrows, which direct the reader's attention to clinically relevant regions and could serve as a valuable supervisory signal in future work.
Apart from ophthalmic applications, PubMed-Ophtha-Annotation could serve a purpose in the improvement and evaluation of figure parsing models, given that it contains manual annotations for 1,885 figures, with 1,534 including image-level annotations.
The dataset can be extended using the automatic pipeline and does not require additional expert annotation, unlike approaches such as MM-Retinal and RetiZero.

While our validation showed high quality across the pipeline, panel identifier localization remained the weakest step, and 6.4\% of the captions could not be processed.
Subcaption assignments can therefore be imprecise for individual panels, which we mitigated through the LLM-based refinement during panel assembly.
The general biomedical literature baseline also differs from the other training data sources in its text encoder, so its results cannot be attributed to the training data alone.

Beyond the comparison itself, these results have implications for how such models could be deployed in clinical research, diagnostics, and patient care.
The k-NN performance of models trained on PubMed-Ophtha indicates that the resulting embedding space captures clinically relevant structure.
This embedding space could potentially be used to retrieve similar cases, match unpaired images and texts using a similarity threshold or automatically filter annotations.
More generally, the improved representations could benefit diagnostic models.
As the improvements are architecture-agnostic, PubMed-Ophtha could be used in any training pipeline.
While we did not investigate multimodal large language models or the effect of finetuning the vision encoder on the evaluation datasets, we hypothesize that the improved vision embeddings likely benefit these settings as well, particularly for rare diseases and conditions.
The strong performance on rare pathologies is particularly relevant for referral in screening settings, where systems trained on categorical labels can only recognize the conditions they were annotated for.
Notably, this advantage originates in the training text but persists in the vision encoder alone, so it remains available in applications that do not involve text at inference time.

Researchers could use the automation to continuously integrate newest clinical findings into their models, keeping them aligned with a field in which new pathologies and imaging modalities emerge over time.
Additionally, almost half of the panels contain retinal imaging, a category comprising ultra-wide field imaging, fluorescein angiography, fundus autofluorescence, and other retinal depictions, for which few public datasets exist.
The hierarchical annotations could further allow pairing related images within a panel, such as a fundus image and the corresponding OCT B-scan, for multimodal training.

While these directions could benefit researchers, clinicians, and patients alike, deployment would require validation on external data and demographic subgroups, as well as calibration of the resulting systems.

We therefore consider domain-specific literature the most effective of the compared sources of textual supervision for fundus images, and release PubMed-Ophtha to make it available for future work.

\section{Data availability}

The datasets are available on Hugging Face Datasets\footnote{\url{https://huggingface.co/datasets/pubmed-ophtha/PubMed-Ophtha}}.
The pubmed\_ophtha.parquet file contains the PubMed-Ophtha dataset in a panel-centric format, and pubmed\_ophtha\_annotation.json contains the ground-truth annotations from the PubMed-Ophtha-Annotation dataset.
Details on the dataset format are available in \hyperref[S3_Table]{Supplementary Tables 3} and \hyperref[S4_Table]{4}, and \nameref{S1_Text}.
All models used to create the dataset and perform the experiments are available on Hugging Face\footnote{Detection models: \url{https://huggingface.co/pubmed-ophtha/detection-models}; Experiments: \url{https://huggingface.co/pubmed-ophtha/experiment-checkpoints}; Downstream use: \url{https://huggingface.co/collections/pubmed-ophtha/pubmed-ophtha-clip-models}}.
We additionally release CLIP models for downstream use that were trained on PubMed-Ophtha with only a 5\% validation set and no test set to make use of the full dataset for training under the MIT license.
Their performance is reported in \nameref{S5_Table}.

\section{Code availability}

The code used to create the dataset and perform the experiments is available on GitHub\footnote{Dataset pipeline: \url{https://github.com/berenslab/pubmed-ophtha}; PDF parser: \url{https://github.com/berenslab/pmo-parser}; CLIP experiments: \url{https://github.com/berenslab/pmo-experiments}} under the MIT license.

\section{Author contributions}

Conceptualization: VH, CE, PB; Methodology: VH, CE, PB; Software: VH; Validation: VH; Data Curation: VH; Writing - Original Draft: VH, PB; Writing - Review  \& Editing: CE; Visualization: VH, PB; Supervision: PB, CE; Funding acquisition: PB

\section{Acknowledgments}
We thank Camila Roa, Sarah Müller, Ifeoma Nwabufo, Jan-Niklas Böhm, Fabio Seel, Samuel Ofosu Mensah, Simone Ebert, Rita González Márquez and Julius Gervelmeyer for annotating PubMed-Ophtha-Annotation.

We thank the Hertie Foundation and the Carl Zeiss Foundation (CZ Nexus: Certification and Foundations of Safe Machine Learning Systems in Healthcare) for funding. PB and CE are members of the Cluster of Excellence 2064 ``Machine Learning -- New Perspectives for Science" funded by the German Research Foundation (DFG).

\section{Competing interests}

CE is a founder and shareholder of Ascléa GmbH.


\newpage
\bibliography{references}

\newpage
\section{Supplementary Information}

\paragraph{Supplementary Fig. 1}
\label{S1_Fig}\label{fig:label-studio-screenshot1}
Labeling interface during the figure and caption localization. The left side contains a screenshot of the PDF page on which the annotators mark the figure and caption bounding boxes.
The right side contains the figure image, caption, and citation from the BIOMEDICA dataset, and an overview of the annotated bounding boxes.
\begin{center}\includegraphics[width=0.9\textwidth]{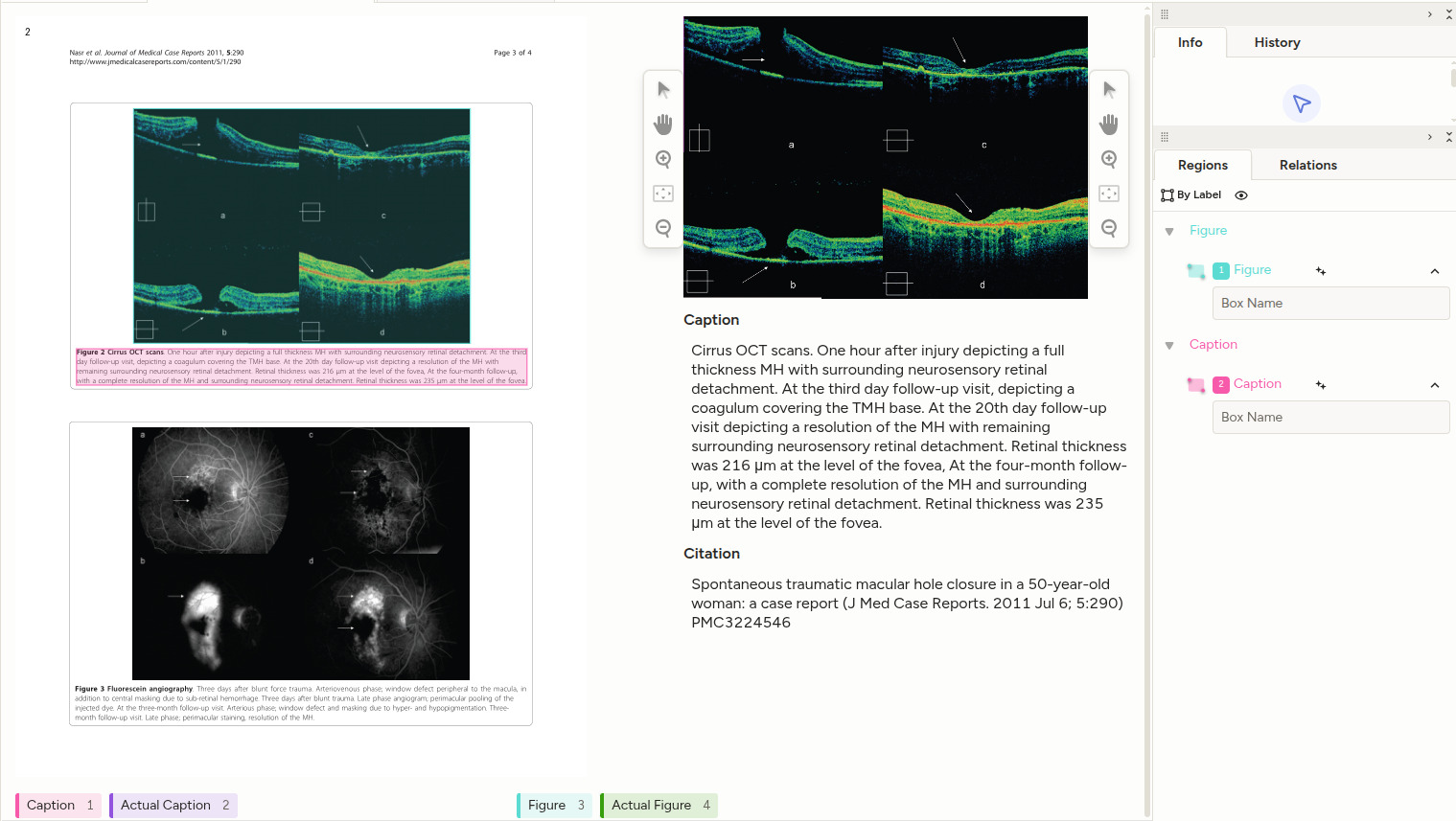}\end{center}
 
\paragraph{Supplementary Fig. 2}
\label{S2_Fig}\label{fig:label-studio-screenshot2}
Labeling interface during panel and image annotation. The left side contains the figure on which the annotators draw the panel, image, or panel identifier bounding boxes.
Right of the figure is the label selection, the figure caption, citation, and keywords.
Note that the annotation type and label fields shown in the interface correspond to the mark status and panel identifier in the final dataset, respectively.
\begin{center}\includegraphics[width=0.9\textwidth]{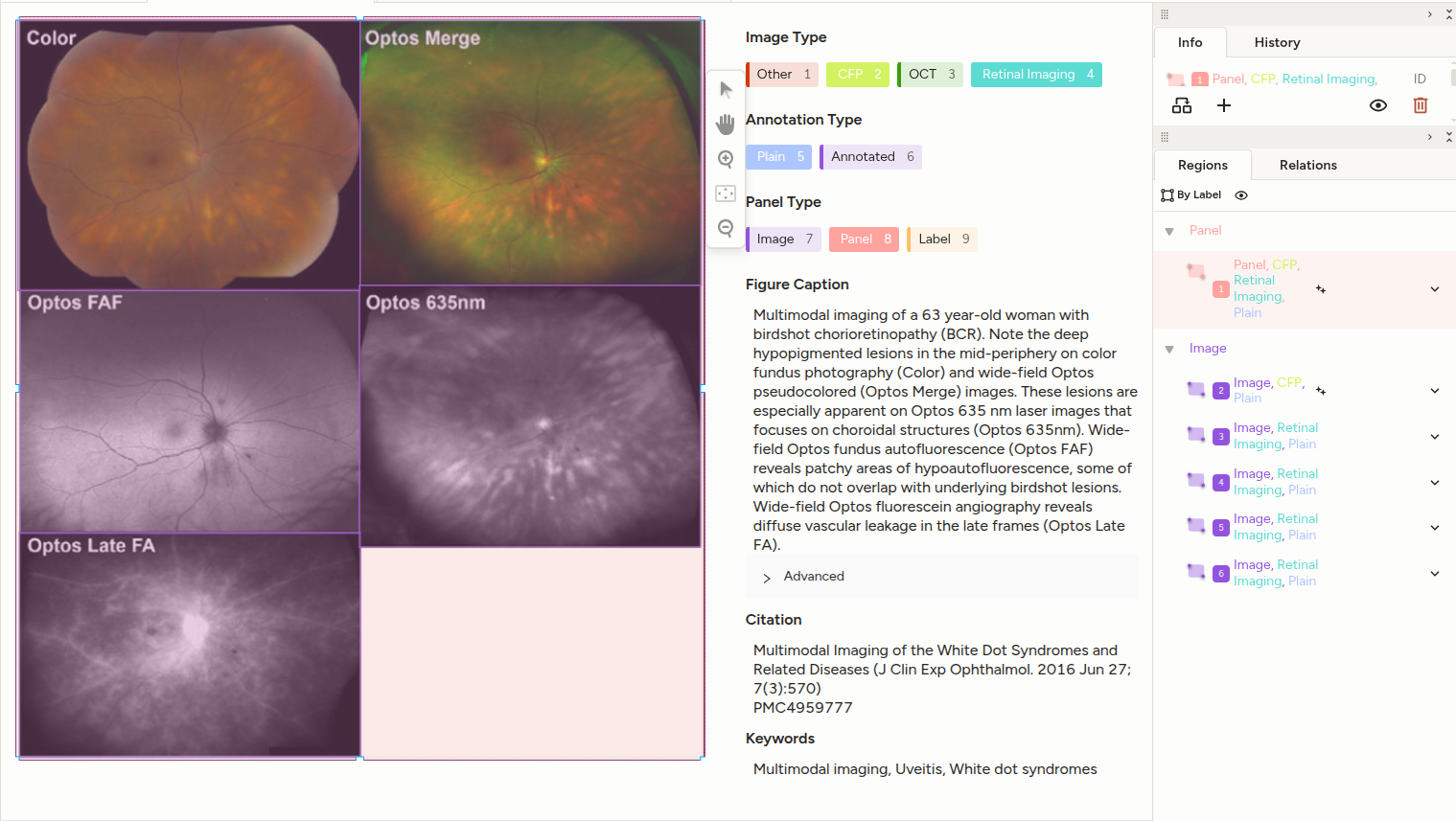}\end{center}

 \newpage
\paragraph{Supplementary Fig. 3}
\label{S3_Fig}\label{fig:label-studio-screenshot3}
Labeling interface during caption annotation. The left side contains the annotated bounding boxes, while the right side contains the label selection, the figure caption, citation, and keywords.
The red arrows associate parts of the caption with panels.
Each panel bounding box has a free-text field for the subcaption (bottom right). 
\begin{center}\includegraphics[width=0.9\textwidth]{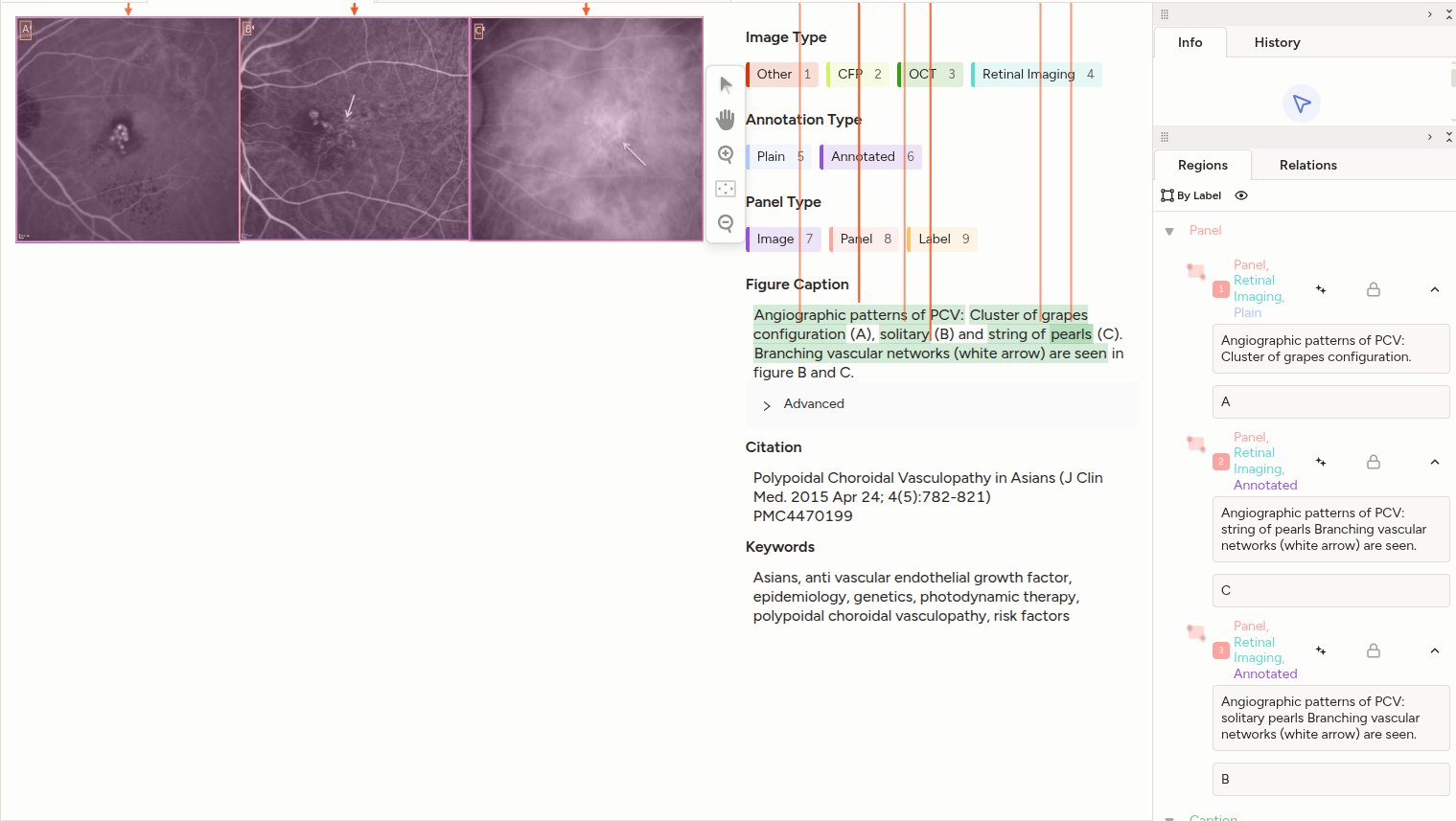}\end{center}

\newpage
\paragraph{Supplementary Fig. 4}\label{fig:many-examples}\label{S4_Fig}
Examples of extracted panels. Each example shows the extracted panel at the top, with a blue square indicating the detected panel identifier position. The detected images, panel identifier text, and subcaption are displayed below it. The symbol in the top right corner of each detected image indicates the mark status, and the text below the image indicates the detected imaging type. 
\begin{center}\includegraphics[width=0.9\textwidth]{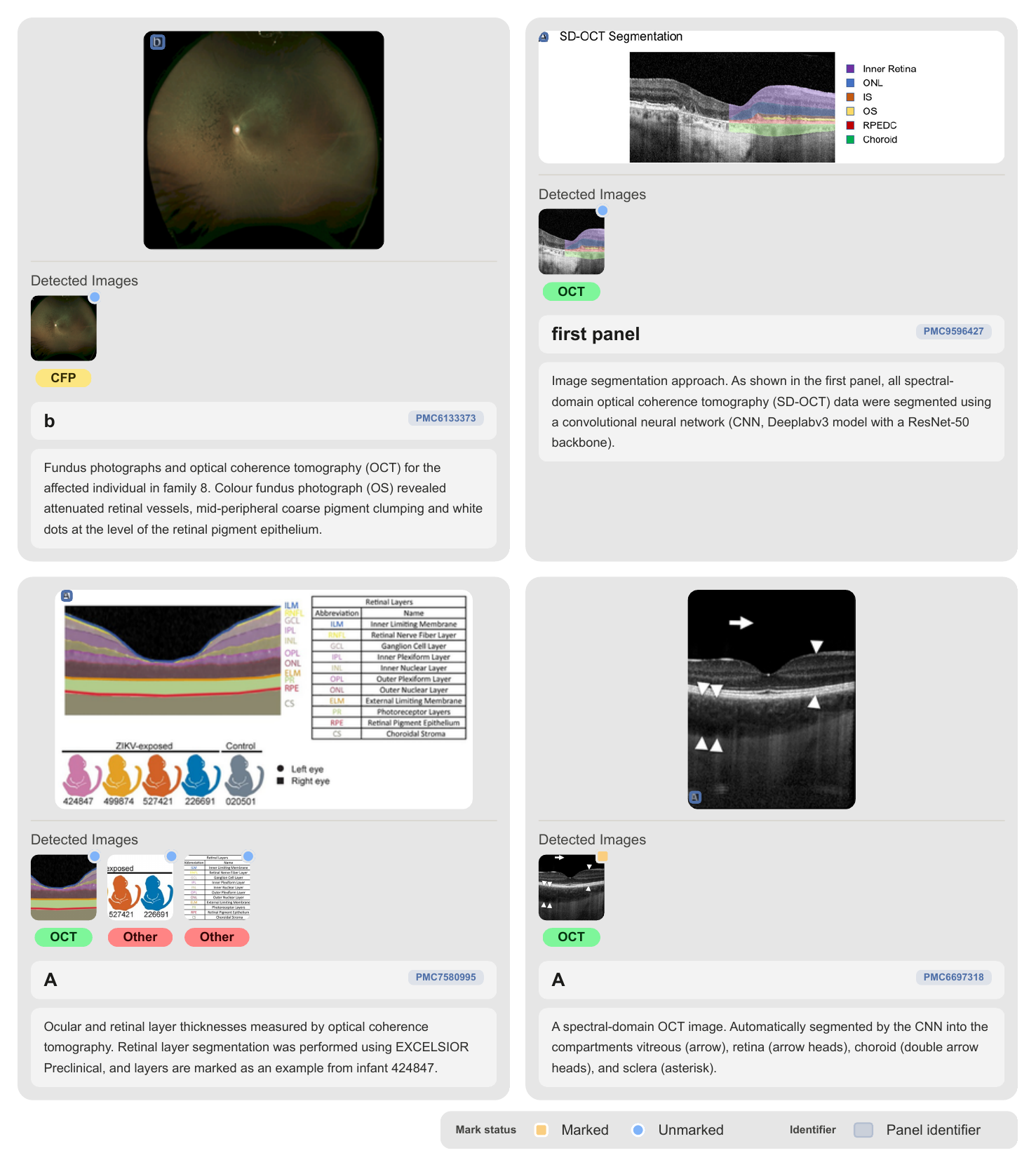}\end{center}

\newpage
\paragraph{Supplementary Table 1}
\label{tab:evaluation_dataset_sizes}\label{S1_Table}
Number of images in the evaluation datasets.

\begin{center}
\footnotesize
\begin{tabular}{lrrr}
\toprule
\textbf{Dataset} & \textbf{Train} & \textbf{Validation} & \textbf{Test} \\
\midrule
APTOS 2019 & 2{,}930  & 366   & 366    \\
EDDFS      & 15{,}602 & 1{,}724 & 11{,}551\\
REFUGE2    & 1{,}200  & 400   & 400     \\
BRSET      & 11{,}060 & 1{,}952 & 3{,}254  \\
RFMiD v1  & 1{,}920  & 640   & 640    \\
RFMiD v2  & 507    & 177   & 170    \\
\bottomrule
\end{tabular}
\end{center}

\paragraph{Supplementary Table 2}
\label{tab:task_groups}\label{S2_Table}
Assignment of evaluation tasks to task groups. Each task is a unique combination of an evaluation dataset and a disease or pathology.
The identifiers correspond to the label columns in the respective evaluation dataset.
AMD: age-related macular degeneration, DR: diabetic retinopathy, HR: hypertensive retinopathy, AH: asteroid hyalosis, MS: macular scars.

\begin{center}
\footnotesize
\begin{tabular}{
    >{\raggedright\arraybackslash}p{0.20\textwidth}
    r
    >{\raggedright\arraybackslash}p{0.18\textwidth}
    >{\raggedright\arraybackslash}p{0.38\textwidth}
}
\toprule
\textbf{Task group} & \textbf{Tasks} & \textbf{Dataset} & \textbf{Label identifier} \\
\midrule
Disease risk & 3 & EDDFS        & \texttt{normal} \\
             &   & RFMiD v1     & \texttt{Disease\_Risk} \\
             &   & RFMiD v2     & \texttt{WNL} \\
\midrule
DR & 6 & APTOS 2019 & \texttt{diagnosis} \\
   &   & BRSET      & \texttt{DR\_ICDR} \\
   &   & BRSET      & \texttt{diabetic\_retinopathy} \\
   &   & EDDFS      & \texttt{DR} \\
   &   & RFMiD v1   & \texttt{DR} \\
   &   & RFMiD v2   & \texttt{DR} \\
\midrule
AMD & 4 & BRSET    & \texttt{amd} \\
    &   & EDDFS    & \texttt{AMD} \\
    &   & RFMiD v1 & \texttt{ARMD} \\
    &   & RFMiD v2 & \texttt{ARMD} \\
\midrule
Glaucoma & 2 & EDDFS   & \texttt{glaucoma} \\
         &   & REFUGE2 & \texttt{glaucoma\_risk} \\
\midrule
Myopia & 4 & BRSET    & \texttt{myopic\_fundus} \\
       &   & EDDFS    & \texttt{myopia} \\
       &   & RFMiD v1 & \texttt{MYA} \\
       &   & RFMiD v2 & \texttt{MYA} \\
\midrule
HR & 3 & BRSET    & \texttt{hypertensive\_retinopathy} \\
   &   & EDDFS    & \texttt{hyper} \\
   &   & RFMiD v2 & \texttt{HTN} \\
\midrule
AH & 2 & RFMiD v1 & \texttt{AH} \\
   &   & RFMiD v2 & \texttt{AH} \\
\midrule
MS & 2 & RFMiD v1 & \texttt{MS} \\
   &   & RFMiD v2 & \texttt{MS} \\
\bottomrule
\end{tabular}
\end{center}

\newpage

\paragraph{Supplementary Table 3}
\label{tab:pubmed-ophtha-column-overview}\label{S3_Table}
Overview of the fields in the pubmed\_ophtha.parquet file.

\begin{center}
    \footnotesize
    \begin{tabular}{
        >{\raggedright\arraybackslash}p{0.26\textwidth}
        >{\raggedright\arraybackslash}p{0.08\textwidth}
        >{\raggedright\arraybackslash}p{0.53\textwidth}
    }
    \hline
    \textbf{Field name} & \textbf{Type} & \textbf{Content} \\
    \hline
    panel\_id & int & Unique ID of the panel \\
    article\_id & int & PMC ID of the article (omitting the PMC in the beginning) \\
    image\_cluster\_id & string & ID of the figure in PMC article \\
    panel\_name & string & Name of the panel inferred from the panel identifier \\
    subcaption\_text & string & Text of the subcaption \\
    subcaption\_segments & list & Indices of the subcaption text in the original caption \\
    panel\_image\_bytes & binary & PNG image containing the panel \\
    position & string & JSON containing positional information.\\
    is\_multi\_page\_figure & bool & True if the figure stretches across multiple pages \\
    in\_text\_mention & list  & In-text mentions of the figure (from BIOMEDICA dataset) \\
    contains\_cfp & bool & True if at least one image in the panel has type CFP \\
    contains\_oct & bool & True if at least one image in the panel has type OCT \\
    contains\_retinal & bool & True if at least one image in the panel has type Retinal Imaging \\
    contains\_other & bool & True if at least one image in the panel has the type Other \\
    contains\_marked & bool & True if at least one image in the panel has a mark \\
    image\_data & string & JSON containing detected images \\
    identifier\_data & string & JSON containing detected panel identifiers \\
    assembly & string & Method of panel assembly \\
    annotation & string & JSON list containing the parts that were humanly annotated (figure, panel, image and caption) \\
    license & string & License under which the paper containing the figure was published \\
    attribution & string & Attribution to the original work \\
    commercial\_use & bool & True if the license allows commercial use \\
    \hline
    \end{tabular}
\end{center}

\paragraph{Supplementary Table 4}
\label{tab:annotated-overview}\label{S4_Table}
Overview of the fields in the pubmed\_ophtha\_annotation.json file.
\begin{center}
    
    \footnotesize
    \begin{tabular}{
        >{\raggedright\arraybackslash}p{0.26\textwidth}
        >{\raggedright\arraybackslash}p{0.08\textwidth}
        >{\raggedright\arraybackslash}p{0.53\textwidth}
    }
    \hline
    \textbf{Field name} & \textbf{Type} & \textbf{Content} \\
    \hline
    article\_id & int & PMC ID of the article \\
    image\_cluster\_id & string & ID of the figure in PMC article \\
    figure\_locations & list & List containing the pages and positions (in pt) of the figure \\
    caption\_locations & list & List containing the pages and positions (in pt) of the captions \\
    panel\_data & list & List containing the annotated panels (if available) with their assigned images, identifiers, and subcaptions \\
    annotation & string & JSON list containing the parts that were humanly annotated (figure, panel, image and caption) \\
    license & string & License under which the paper containing the figure was published \\
    attribution & string & Attribution to the original work \\
    \hline
    \end{tabular}
\end{center}

\newpage

\paragraph{Supplementary Table 5}
\label{tab:downstream-clip-models}\label{S5_Table}
Performance of the released CLIP models for downstream use. Each score is the mean AUROC (\%, $\uparrow$) across all tasks that fall into the task group; the average column indicates the mean score across all 110 tasks.
Training followed the same procedure as the PubMed-Ophtha models in the main text (see \nameref{subsec:model-training}), except that the models were trained on a 95/5 train/validation split without a held-out test split, on a single seed.
They are therefore not part of the comparison reported in the main text.
DR: diabetic retinopathy, AMD: age-related macular degeneration, HR: hypertensive retinopathy, AH: asteroid hyalosis, MS: macular scars.

\begin{center}
\footnotesize
\setlength{\tabcolsep}{4pt}
\resizebox{\ifdim\width>\linewidth\linewidth\else\width\fi}{!}{%
\begin{tabular}{llccccccccc}
\toprule
\textbf{Setting} & \textbf{Backbone} & \textbf{Disease risk} & \textbf{DR} & \textbf{AMD} & \textbf{Glaucoma} & \textbf{Myopia} & \textbf{HR} & \textbf{AH} & \textbf{MS} & \textbf{Average} \\
\midrule
k-NN           & ViT-B/32 & 92.39 & 88.65 & 78.49 & 75.49 & 97.18 & 75.19 & 89.62 & 63.12 & 75.90 \\
               & ViT-B/16 & 93.60 & 92.15 & 79.96 & 78.03 & 97.24 & 76.76 & 94.46 & 77.31 & 77.57 \\
               & ViT-L/14 & 94.02 & 93.10 & 80.21 & 79.54 & 97.06 & 77.60 & 89.86 & 75.00 & 78.04 \\
\midrule
Zero-shot      & ViT-B/32 & 80.30 & 73.81 & 74.58 & 59.44 & 89.29 & 53.90 & 88.79 & 80.99 & 72.12 \\
               & ViT-B/16 & 85.95 & 77.50 & 77.51 & 60.48 & 80.58 & 67.43 & 93.23 & 78.07 & 73.39 \\
               & ViT-L/14 & 84.32 & 77.69 & 81.15 & 58.73 & 86.69 & 50.33 & 85.19 & 80.91 & 74.78 \\
\midrule
Linear probing & ViT-B/32 & 92.93 & 91.78 & 93.17 & 86.75 & 99.20 & 82.82 & 98.99 & 83.45 & 87.69 \\
               & ViT-B/16 & 94.33 & 95.01 & 94.40 & 88.22 & 98.86 & 82.33 & 97.43 & 90.46 & 89.45 \\
               & ViT-L/14 & 94.46 & 96.17 & 94.89 & 86.95 & 98.86 & 83.57 & 99.01 & 87.41 & 89.83 \\
\bottomrule
\end{tabular}
}
\end{center}

\newpage

\paragraph{Supplementary Note 1}\label{S1_Text}
We provide the PubMed-Ophtha dataset in a panel-centric format in the pubmed\_ophtha.parquet file to facilitate the training of VLMs.
In the file, each row represents a panel from a figure with a unique panel\_id and a set of features, including the panel as binary image, the subcaption text and binary imaging type and mark status indicators.
The full list of columns in the file and their descriptions can be found in \nameref{S3_Table}.
The \textit{position} column contains a JSON-serialized dictionary with four entries: \textit{predicted\_box}, the panel bounding box as predicted by the panel detection model; \textit{content\_box}, the final bounding box encompassing all panel contents including associated images and panel identifiers; both normalized to figure dimensions (left, top, right, bottom in the range $[0, 1]$); \textit{figure\_page\_coordinates}, the figure position in PDF page coordinates; and \textit{prediction\_score}, the detection confidence of the predicted panel bounding box.
The \textit{image\_data} and \textit{identifier\_data} columns contain JSON-serialized lists of the images and panel identifiers associated with the panel.
Each entry stores its bounding box in the \textit{position} field as a JSON-serialized dictionary containing \textit{predicted\_box} -- coordinates normalized relative to the figure dimensions (left, top, right, bottom in the range $[0, 1]$), and a \textit{prediction\_score}.
Each image additionally has an \textit{imaging\_type} field with a label and prediction score, a \textit{mark\_status} field indicating whether the image contains marks, and flags indicating whether it originates from ground-truth annotation (\textit{is\_gt}) and whether it is used to calculate the panel content box (\textit{included\_in\_content\_box}).
Each panel identifier entry similarly contains \textit{is\_gt}, \textit{included\_in\_content\_box}, and an \textit{origin\_identifier} field linking back to the detection from which it was derived.
Note that neither the PMC ID nor the image cluster ID alone uniquely identifies a figure, as only their combination does.
\nameref{S4_Fig} shows several examples from the dataset.

The second file, pubmed\_ophtha\_annotation.json, is the PubMed-Ophtha-Annotation dataset, where each entry corresponds to one figure.
Each figure is uniquely identified by the combination of its \textit{article\_id} and \textit{image\_cluster\_id} and has a list of figure and caption locations and a list of panels.
The locations of the figures and captions are provided as a list because the figures can spread across multiple pages, while the positions of the panels, images, and panel identifiers are normalized in figure space.
The fields in the JSON file are described in \nameref{S4_Table}.
For some figures, features such as the panel images and subcaptions are not available upon download due to license restrictions.
These fields can be populated using the scripts in the repository.
Additionally, the repository contains examples for data loading and usage in Python\footnote{\url{https://github.com/berenslab/pubmed-ophtha/tree/main/examples}}.

\end{document}